\documentclass[sigconf]{acmart}
\AtBeginDocument{%
  \providecommand\BibTeX{{%
    \normalfont B\kern-0.5em{\scshape i\kern-0.25em b}\kern-0.8em\TeX}}}

\usepackage{times}
\usepackage{latexsym}
\usepackage[T1]{fontenc}
\usepackage[utf8]{inputenc}
\usepackage{microtype}
\newcommand{\hide}[1]{} 
\usepackage{xspace}
\usepackage[ruled,vlined]{algorithm2e}
\usepackage{graphicx}
\usepackage{multirow}
\usepackage{subfigure}
\usepackage{amsmath}
\usepackage{bbm}
\usepackage{textcomp}
\usepackage{dsfont}
\usepackage{amsfonts}       
\newcommand*{\eg}{\emph{e.g.},\@\xspace}
\newcommand*{\ie}{\emph{i.e.},\@\xspace}
\newcommand*{\etc}{\emph{etc.}\@\xspace}

\DeclareMathOperator*{\argmax}{arg\,max}
\DeclareMathOperator*{\argmin}{arg\,min}

\newcommand{\norm}[1]{\left\lVert#1\right\rVert}

\newcommand{\il}[1]{\textsf{\textbf{\color{magenta}{\small{[IL: #1]}}}}}

\newcommand*{\modelname}{SAUCE\@\xspace}

\copyrightyear{2021}
\acmYear{2021}
\setcopyright{acmlicensed}\acmConference[CIKM '21]{Proceedings of the 30th ACM International Conference on Information and Knowledge Management}{November 1--5, 2021}{Virtual Event, QLD, Australia}
\acmBooktitle{Proceedings of the 30th ACM International Conference on Information and Knowledge Management (CIKM '21), November 1--5, 2021, Virtual Event, QLD, Australia}
\acmPrice{15.00}
\acmDOI{10.1145/3459637.3481950}
\acmISBN{978-1-4503-8446-9/21/11}

\begin{document}
\settopmatter{printacmref=true,authorsperrow=4}

\title{\modelname: Truncated Sparse Document Signature Bit-Vectors for Fast Web-Scale Corpus Expansion}
\fancyhead{} 
\author{Muntasir Wahed}
\affiliation{
  \institution{Virginia Tech}
}
\email{mwahed@vt.edu}

\author{Daniel Gruhl}
\affiliation{
  \institution{IBM Research Almaden}
}
\email{dgruhl@us.ibm.com}

\author{Alfredo Alba}
\affiliation{
  \institution{IBM Research Almaden}
}
\email{aalba@us.ibm.com}

\author{Anna Lisa Gentile}
\affiliation{
  \institution{IBM Research Almaden}
}
\email{annalisa.gentile@ibm.com}

\author{Petar Ristoski}
\affiliation{%
  \institution{eBay Inc}
}
\email{pristoski@ebay.com}

\author{Chad Deluca}
\affiliation{%
  \institution{IBM Research Almaden}
}
\email{delucac@us.ibm.com}

\author{Steve Welch}
\affiliation{
\institution{IBM Research Almaden}
}
\email{welchs@us.ibm.com}

\author{Ismini Lourentzou}
\affiliation{
  \institution{Virginia Tech}
}
\email{ilourentzou@vt.edu}
\renewcommand{\shortauthors}{Wahed, et al.}

\begin{abstract}
Recent advances in text representation have shown that training on large amounts of text is crucial for natural language understanding. However, models trained without predefined notions of topical interest typically require careful fine-tuning when transferred to specialized domains. When a sufficient amount of within-domain text may not be available, expanding a seed corpus of relevant documents from large-scale web data poses several challenges. First, corpus expansion requires scoring and ranking each document in the collection, an operation that can quickly become computationally expensive as the web corpora size grows. Relying on dense vector spaces and pairwise similarity adds to the computational expense. Secondly, as the domain concept becomes more nuanced, capturing the long tail of domain-specific rare terms becomes non-trivial, especially under limited seed corpora scenarios. 

In this paper, we consider the problem of fast approximate corpus expansion given a small seed corpus with a few relevant documents as a query, with the goal of capturing the long tail of a domain-specific set of concept terms. To efficiently collect large-scale domain-specific corpora with limited relevance feedback, we propose a novel truncated sparse document bit-vector representation, termed Signature Assisted Unsupervised Corpus Expansion (\modelname). Experimental results show that \modelname can reduce the computational burden while ensuring high within-domain lexical coverage.
\end{abstract}


\maketitle

\section{Introduction}\label{sec:intro}

Unsupervised machine learning approaches have been increasingly successful in discovering patterns from text, image and trajectory data \cite{pagliardini2018unsupervised,lample2018unsupervised,sharma2019dynamics,schmitt2020unsupervised}. Yet, results heavily depend on the quality of the provided training data. As demand for larger text collections grows, \eg language models trained on terabytes of text data, the web remains the main large-scale and dynamic resource of diverse text \cite{devlin2019bert,liu2019roberta, brown2020language}. 
However, such text collection is typically performed without any predefined notions of topic interest, and language models trained on general text corpora may perform well on common tasks but require careful priming when transferred to specialized domains, such as bio-medical, financial and legal domains~\cite{lee2020biobert,araci2019finbert,ma2019domain}. 

Recent work emphasizes the importance of domain-specific data, showing improvements over general language models \cite{gururangan2020don,edwards2020go,komatsuzaki2020current}. Document filtering and unsupervised retrieval methods have emerged as a solution, each facing different challenges. Generally, filtering leads to computational resources being unnecessarily wasted, as large amounts of data have to be downloaded, yet most of them will be discarded. On the other hand, retrieval methods used for corpora collection rely on computing vector-space similarity between a single query and each document in the collection. However, examining such pairwise distances involves considerably more computational resources as the corpora size grows non-linearly.

Another important factor is that term frequency follows a Zipfian distribution, implying that a corpus collection method should properly capture the long tail of domain-specific rare terms. Finally, the notion of topical interest depends on the task at hand and subsequently on the domain expert that provides relevance feedback. In a typical search, this feedback is noisy and implicit. In contrast, when collecting training data for domain-specific fine-tuning, relying on click-logs from general crowds is not ideal for high-quality priming examples. Therefore, there is an inherent trade-off between minimizing the subject matter expert (SME) supervision and maximizing the coverage of the domain at hand.

In this paper, we investigate several document representation techniques, ranging from bag-of-words to pre-trained contextualized language models. Accordingly, based on our analysis on the trade-off between efficiency and accuracy, we present a corpus expansion method, termed \textbf{S}ignature \textbf{A}ssisted \textbf{U}nsupervised \textbf{C}orpus \textbf{E}xpansion (\textbf{\modelname}), that can address the aforementioned challenges. Specifically, \modelname addresses the real-world problem of extending domain-specific corpora by retrieving relevant documents from large-scale web collections.

In contrast with previous domain-specific corpora expansion approaches, our method extends bit-vector signature methods \cite{faloutsos1990signature,geva2011topsig}. We leverage the fact that only a small number of low-frequency terms captures the topic of a document \cite{klein2008revisiting}, to purposefully generate ``weak'' signatures that capture the domain concept loosely enough to pick up variant terminologies, essentially building extremely sparse document vector representations. Experimental results showcase that \modelname can reduce the query execution time by $78\%$ and the per-document memory footprint by $24\%$. Most importantly, \modelname allows for broader coverage of the domain vocabulary, as it can retrieve up to $6.8\%$ more within-domain terms when compared to baselines.

The following are the major contributions of this work: 
\begin{itemize}
    \item We propose a novel efficient large-scale corpus expansion framework, termed \modelname, for expanding or creating within-domain collections from a small seed corpus of relevant documents. 
    \item We present a method for constructing document signatures, \ie truncated sparse bit-vector representations, in an unsupervised fashion. The proposed representation is easily parallelizable and adaptive to new streams of text.
    \item Experimental results show that our proposed document signatures effectively reduce the retrieval time from petabytes of web documents while maintaining broader coverage of the long tail domain vocabulary distribution.
\end{itemize}

The remainder of this paper is structured as follows. We describe related work in Section \ref{sec:related}. In Section \ref{sec:problem} we formally explain the task and outline its importance, while in Section \ref{sec:method} we introduce our approach. We present an experimental evaluation in Section \ref{sec:exp}, followed details on system deployment in Section \ref{sec:deploy}, and conclusions and future work in Section \ref{sec:conclusion}.

\section{Related Work}\label{sec:related}

\subsection{Concept Expansion}
Given a set of seed entities representing a concept, concept expansion methods, otherwise termed set expansion, query a web data source with the goal of expanding the set with other entities belonging to the same conceptual class. For example, a seed query may contain terms such as ``Prague'', ``Athens'', ``Paris'', ``Rome'' as seed entities and a concept expansion method would derive other entities (\eg, ``Munich'', ``Vienna'', ``London'', \etc) that capture the concept set of cities in Europe. Albeit there is a significant body of work on concept expansion, most methods are non-contextual, \ie using the seed entities as query does not provide any context in which the entities might appear. Here, we briefly summarize some of the proposed methods. 

\citet{ghahramani2005bayesian} formulate concept expansion as a Bayesian inference problem and evaluate the marginal probability of each item belonging to the same cluster as the query items. Other early approaches rely on co-occurrence statistics or graph-based random walks to rank candidates entities \cite{sarmento2007more,Wang2008}. \citet{Pantel2009} propose a scalable MapReduce framework for calculating co-occurrence statistics and pairwise similarities and evaluate on the task of automatic set expansion. \citet{He2011} propose a metric to measure the quality of the expanded set that is used as the basis for two proposed iterative thresholding algorithms that rank candidate entities. In contrast with previous pattern matching approaches for extracting candidate entities, \citet{Chen2016} leverage webpage structural and textual information to build page-specific extractors. Their method requires distantly supervised negative examples that correspond to mutually exclusive concepts useful for disambiguation and for mitigating semantic drift \cite{etzioni2005unsupervised,carlson2010toward,curran2007minimising}. \citet{Wang2015} utilize the seed entities to retrieve web tables, construct bipartite graphs that capture <entity,table> relationships, and rank candidates with an exclusivity-based probabilistic method. 
\citet{Yu2019} leverage both lexical features and distributed representations to expand input seed entities with semantically related entities that frequently share context with seed entities. The authors, later on, propose a listwise neural re-ranking model, trained using entity interaction features, that jointly learns two semantic and exact matching channels \cite{Yu2020}.
A limitation of the aforementioned methods is that, due to the lack of contextual information when performing a query, they rely on external resources for concept disambiguation.

Such contextualized concept expansion tasks have only been recently studied. \citet{Han2019} propose a neural attention-based architecture for Context-Aware Semantic Expansion (CASE), where the seed entity set query is replaced with a sentence query that provides more context, and candidate entities are retrieved based on whether they fit the context well. However, finding or generating one representative sentence is not always straightforward. Moreover, our experiments show that such query-based methods do not model the long tail well and remain insufficient in terms of query execution time. Both document retrieval and corpus expansion methods allow the subject matter expert (SME) to provide more context in the form of relevant documents. 

\subsection{Document Retrieval}
Ad-hoc document retrieval is framed as the task of producing an ideal ranking of documents in a corpus  $\mathcal{C} =\{\mathcal{D}_i\}_{i=1}^N$, conditioned on a short query $\mathcal{Q}$, and is typically evaluated according to some standard metric such as average precision, recall or normalized discounted cumulative gain \cite{manning2008introduction}. Dominant approaches involve cascade architectures, where standard bag-of-words term-matching techniques are first employed towards creating the initial list of candidate documents, followed by a neural re-ranker, typically a language model that performs additional rescoring \cite{yang2019simple}. 

The success of adapting a neural language model to the document retrieval task or utilizing supervised learning-to-rank models heavily depends on the quality and quantity of the training data, which often include large amounts of relevance judgments on document or passage level \cite{yang2019simple,huang2013learning}. Moreover, such models require powerful computing infrastructures and can take a long time to train, making them impractical for adaptive online scenarios. Thus, in this work, we investigate unsupervised domain-specific corpus collection, taking into account the scalability of the method, the limited relevance feedback available and the coverage of the domain terminology.

Unsupervised retrieval typically involves scoring documents with some form of vector similarity between the query and each document, \eg TF-IDF or vectors originating from pre-trained embedding models publicly available online \cite{salton1971smart,gysel2018neural}. The performance of such pairwise similarity scoring depends on the dimensionality of the vector space. On one hand, TF-IDF requires document frequency, which can be slow to compute and cannot be parallelized easily or used for streams of text. On the other hand, the use of black-box pre-trained embedding models heavily depends on the quality of model training, but information on the quality or the within-domain evaluation is not always available to the end-user. Finally, pairwise similarity involves $O(dNQ)$  operations, where  $d$ is the dimensionality of the vector space, $N$ is the number of documents in the corpus and $Q$ is the number of queries. In this work, we show that the aforementioned approaches are computationally expensive and do not capture the long tail distribution of highly specialized domains.

Leveraging approximate nearest neighbor search has surfaced as a viable solution,
where results are not exact and hence some vectors with high similarity are missed \cite{JDH17,malkov2018efficient,baranchuk2018revisiting,xiong2020approximate}. Another approach is to impose sparsity in the dimensions, for example, dimensionality reduction, hashing and feature selection methods can be used to scale on large text collections \cite{pearson1901liii,scholkopf1998nonlinear,guyon2003introduction,xu2019review}. Such methods are not designed for streaming text retrieval systems. In contrast, document signatures are an efficient alternative, but they pose several challenges \cite{goodwin2017bitfunnel}. Document signatures based on probabilistic data structures introduce noise and degrade retrieval quality, as irrelevant pages are added to the results.

Finally, several works deal with the task of corpus expansion, but rely on supervision \cite{remus2016domain,chung2017sentence} or incorporate rules and constraints designed for specific tasks, \eg statistical machine translation \cite{gao2011corpus}, text simplification \cite{katsuta2019improving} and entity set expansion \cite{huang2020guiding}. In this work, we explore signature-based retrieval methods as a computationally efficient alternative for corpus expansion such that we better capture the long tail in domain-specific applications.

\subsection{Corpus Expansion}
Corpus expansion methods typically are used for collecting or expanding parallel corpora for downstream applications.
\citet{Gao2011} frame the task as a variation of paraphrasing, where parallel sentences convey different information, as opposed to traditional paraphrasing that retains the same information in between original and parallel sentences. The authors use semantic role labeling to extract candidate substitution rules for replacing phrases in existing sentences, and a classifier that determines whether a generated sentence is syntactically and semantically well-formed. \citet{Qiu2014} propose a redundancy-based corpus expansion method for Chinese word segmentation that automatically expands the training corpus for uncertain segmentations by searching the web. \citet{Al-Natsheh2017} extract semantic features from a set of scientific articles of interest in order to expand the set with other related academic papers. \citet{Katsuta2019} construct a web corpus to improve sentence simplification. \citet{Remus2016} present a focused crawling corpus expansion method that estimates the relatedness of documents using n-gram language models, while \citet{Chung2017} propose a sentence-chain-based Seq2Seq model trained for corpus expansion. Additionally, there exist works that utilize web crawling to build contextual graph models for seed expansion \cite{remus2016domain, CHEN201833, Abulaish2020}. However, previous work either relies on links between documents to traverse the web graph (which may not always be available) or is designed with specific tasks in mind. To the best of our knowledge, none of the previous works focus on efficiency and domain lexical coverage. Our approach fundamentally differs in that we develop a general, unsupervised and scalable method that can rapidly produce good results from an \textit{unstructured} web-scale document collection, taking into account the trade-off between corpus expansion time and precision.
\begin{figure*}[t!]
\centering
\includegraphics[width=0.9\textwidth]{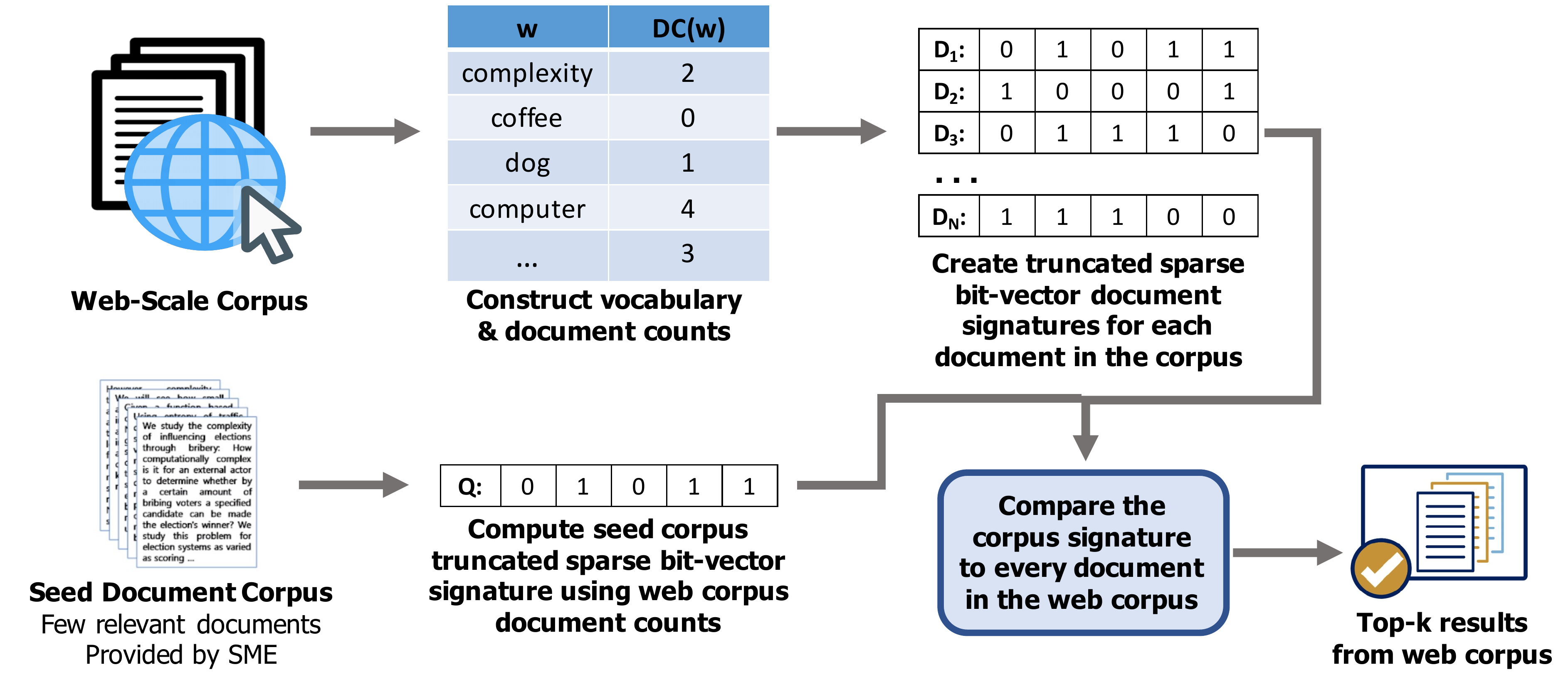}
\caption{Overview of \modelname. Truncated sparse bit-vector document signatures provide an efficient representation for scalable pairwise similarity computation from petabytes of web documents.} 
\label{fig:overview}
\end{figure*}
\section{Problem Statement}\label{sec:problem}
Given a large text collection $\mathcal{C}$, we focus on the problem of identifying a set of $k$ documents that match a set of relevant documents used as the search query. Let corpus $\mathcal{C}$, comprised of a set of $N$ text documents $\mathcal{C} = \{\mathcal{D}_1, \ldots, \mathcal{D}_N \}$, where each document consists of a set of terms, \ie $\mathcal{D}_i = \{w_1^{(\mathcal{D}_i)}, \ldots, w_{T_i}^{(\mathcal{D}_i)}\}$, and $T_i$ is the document length for $\mathcal{D}_i$, \ie the document length can vary. Define a seed  query $\mathcal{Q} = \{q_1, \ldots, q_{T}\}$, where each item $q_j$ is a document that is deemed within-domain from the subject matter expert (SME). In other words, corpus expansion is primed with a small seed corpus $\mathcal{Q}$. Then the goal is to create a scoring function $Score(\mathcal{Q},\mathcal{D}_i)$ that ranks documents based on the similarity with the given query, and then is used to retrieve the top-$k$ documents based on their scores, \ie
\begin{equation}
\argmax\limits_{R \subset \mathcal{C},|R|=k} \sum\limits_{\mathcal{D}_i \in R} Score(\mathcal{Q},\mathcal{D}_i)
\end{equation}

In traditional ad-hoc retrieval, a popular scoring mechanism is TF-IDF, where the document score is computed as the product of the term frequency and the inverse document frequency, for terms present in both document $\mathcal{D}_i$ and the query $\mathcal{Q}$ \cite{salton1971smart}.
Recent works represent text with dense vector representations extracted from language models that capture term co-occurrence statistics, and compute document similarity with pairwise distance metrics such as cosine \cite{gysel2018neural}:  
\begin{align}
\vec{\mathcal{D}_i} = \frac{1}{T_i} \sum\limits_{m=1}^{T_i} \vec{w}_m^{(\mathcal{D}_i)}\\
\vec{\mathcal{Q}} = \frac{1}{T} \sum\limits_{j=1}^{T} \left( \frac{1}{T_{q_j}} \sum\limits_{m=1}^{T_{q_j}} \vec{w}_m^{(q_j)} \right)\\ 
Score(\vec{\mathcal{Q}},\vec{\mathcal{D}_i}) = \frac{\vec{\mathcal{Q}} \cdot \vec{\mathcal{D}_i}}{ \norm{\vec{\mathcal{Q}}} \cdot \norm{\vec{\mathcal{D}_i}}},
\end{align}
where $\vec{\mathcal{D}_i}$ is the vector representation of document $\mathcal{D}_i$, represented by averaging its constituent word vector representations $\vec{w}_m^{(\mathcal{D}_i)}, m=1, \ldots, T_i$. Similarly, the seed query vector representation $\vec{\mathcal{Q}}$ is constructed by averaging the representations of each query $q_j$, which in turn are mean representations of $q_j$'s word vector representations $\vec{w}_m^{(q_j)}, m=1, \ldots, T_{q_j}$.

However, pairwise similarity on dense vectors is a computationally expensive procedure. Moreover, learning representations add to the overall cost. Another option would be to use existing pre-trained embedding models, but these are treated as black-box tools, \ie how useful such a model is for a given task might heavily depend on the quality of training; information that is not always available to the end-user. To this end, we present a novel bit-vector representation with the advantage of being easily parallelizable and adaptive to new streams of text.

\section{The \modelname Method}\label{sec:method}
Document signatures have a longstanding history in information retrieval \cite{geva2011topsig,faloutsos1990signature,chappell2013efficient} but have been less prevalent amidst the deep learning current trends in language modeling. In this work, we argue that signature-based retrieval methods are computationally efficient for corpus expansion and can better capture the long tail in domain-specific applications.

In contrast with previous document signature approaches, we present a method that relies on sparse truncated binary signatures as a document representation that allows for fast approximate corpus expansion. Unlike typical linguistic signatures, we are purposefully generating ``weak''
signatures that capture the domain concept loosely enough to pick up variant terminologies, under the assumption that only a small number of low-frequency terms capture the ``aboutness'' of a document \cite{klein2008revisiting}.

We compare our method with several baselines, providing key insights on the trade-offs and challenges that large-scale corpus expansion systems face. A pictorial overview of the proposed method can be found in Figure \ref{fig:overview}.

\subsection{Document Signatures}

A lexical signature is a small set of terms that is derived from a document and represents the most significant terms of its textual content \cite{klein2008revisiting}. Here, the signature of a document is constructed as a sequence of bits that represents terms found in the document. To reduce the dimensionality, we additionally clip the signatures based on document frequency thresholds. 

More specifically, given a document corpus $\mathcal{C} = \{\mathcal{D}_1, \ldots, \mathcal{D}_N \}$, where each document is of the form $\mathcal{D}_i = \{w_1^{(\mathcal{D}_i)}, \ldots, w_{T_i}^{(\mathcal{D}_i)}\}$, we first construct a vocabulary containing all unique terms,
\begin{equation}
    V = \bigcup\limits_{i=1}^{N} \mathcal{D}_i = \{\tilde{w}_1, \ldots, \tilde{w}_m, \ldots, \tilde{w}_{|V|}\}.
\end{equation}

To create a bit-vector signature $v_i$ for a document $\mathcal{D}_i$, a straight-forward option would be to assign zero to terms that do not appear in text and 1 otherwise, \ie
\begin{equation}
    v(\mathcal{D}_i) = \left[ v_1^{(\mathcal{D}_i)}, \ldots, v_m^{(\mathcal{D}_i)}, \ldots, v_{|V|}^{(\mathcal{D}_i)}\right]
\end{equation}
such that $v_m^{(\mathcal{D}_i)} = \mathbbm{1} \{ \tilde{w}_m \in \mathcal{D}_i \}$.

However, depending on the vocabulary size $|V|$, such vectors can be very high-dimensional, making the computation of pairwise distances expensive. Intuitively, very common terms that appear in a lot of documents will not be as informative, as they do not capture the document topic. On the other hand, very rare terms will not provide wide coverage of the domain when searching for relevant documents. Thus, we propose two threshold strategies to reduce vector dimensionality. 
To this end, we compute document counts
\begin{equation}
DC(\tilde{w}_m) = \sum\limits_{i=1}^{N} \mathbbm{1} \{\tilde{w}_m \in \mathcal{D}_i\}, \forall m=1, \ldots, |V|.
\end{equation}
Then, we reduce the dimensionality of the vector based on whether a term in the vocabulary appears in \textit{at least} $k_1$ documents
\begin{equation}
d = |V| - \sum\limits_{m=1}^{V}  \mathbbm{1} \{DC(\tilde{w}_m) < k_1\}.
\end{equation}.

Note that we use this clipping to construct document vectors in a dimension $d < |V|$, but we keep the terms in the vocabulary, as well as their respective documents counts. This is essential for updating the vectors when a new document appears. The signature bit-vector of document $\mathcal{D}_i \in \mathcal{C}$ is 
\begin{equation}
    v(\mathcal{D}_i) = \left[ v_1^{(\mathcal{D}_i)}, \ldots, v_m^{(\mathcal{D}_i)}, \ldots, v_{d}^{(\mathcal{D}_i)}\right] 
\end{equation}
where 
$v_m^{(\mathcal{D}_i)} =  \mathbbm{1} \{ \tilde{w}_{j_{m}} \in \mathcal{D}_i \}$ is the binary indicator of $\tilde{w}_{j_{m}}, m = 1, \ldots, d$, and $J=\{j_1, \ldots, j_d\}$  set of ordered indices for the remaining terms in the vocabulary, \ie we sort vocabulary words with respect to document counts. 

Finally, we impose additional sparsity to each document vector by removing very common terms. More specifically, in each document $\mathcal{D}_i$, we keep the $k_2$ \textit{least frequent} terms, \ie 
\begin{align}
 v_r(\mathcal{D}_i) = \mathbbm{1} \{r \in R_i\}, \text{where } 
 R_i = \argmin\limits_{R \subset J,|R|=k_2} \sum\limits_{r \in R} DC(\tilde{w}_r).
\end{align}
    
In other words, the set $R_i$ contains the indices of $k_2$ terms found in $\mathcal{D}_i$, with the lowest document count. Thus, all document vectors have at most $k_2$ non-zero bits, making this strategy an efficient approximation. We term our proposed method Signature Assisted Unsupervised Corpus Expansion (\modelname).

\subsection{Updating on New Documents}
When a new document arrives, we update the vocabulary from $V$ to $\hat{V}$ and the document counts. The signature dimension is updated as follows:
\begin{equation}
    \hat{d} = |\hat{V}| - \sum\limits_{m=1}^{|\hat{V}|} \mathbbm{1} \{ DC(\tilde{w}_m) < k_1\}
\end{equation}
For brevity, the unique terms notation is preserved. Note that $|\hat{V}|$ increases as new terms arrive. The dimensionality of the document vectors will increase when there is a need for appending $\hat{d} - d$ terms based on the $(k_1, k_2)$ updated thresholds.
Intuitively, if a new unseen term arrives (that was not part of the initial vocabulary), it will be a rare term that appears in less than $k_1$ documents, and hence document vectors need not be updated. A term that is already in the vocabulary set will result in vector changes only if it surpasses the $k_1$ threshold. At the same time, since we keep the $k_2$ least frequent terms, the effective number of bits required remains the same. We present the pseudocode for the construction of the initial signatures and the pseudocode for the adaptive updates in Algorithms \ref{alg:construct} and \ref{alg:update}, respectively.
\begin{algorithm}[t!]
\DontPrintSemicolon
\textbf{Input:}{\\$\mathcal{C} =\{\mathcal{D}_1, \ldots, \mathcal{D}_N \}$: set of documents\\
$\mathcal{D}_i =  \{w_1^{(\mathcal{D}_i)}, \ldots, w_{T_i}^{(\mathcal{D}_i)}\}, i=1,\ldots,N$ \\
$k_1, k_2$: document thresholds \\}
\Begin{
    Construct vocabulary\\ $V = \bigcup\limits_{i=1}^{N} \mathcal{D}_i = \{\tilde{w}_1, \ldots, \tilde{w}_m, \ldots, \tilde{w}_{|V|}\} $\\
    Compute document counts \\
    \For{$m=1, \ldots, |V|$}
    {
    $DC(\tilde{w}_m) = \sum\limits_{i=1}^{N} \mathbbm{1} \{\tilde{w}_m \in \mathcal{D}_i\}$
    }
    Define signature dimension $d = |V| - \sum\limits_{m=1}^{V} \mathbbm{1} \{DC(\tilde{w}_m) < k_1\}$\\
    Order words in decreasing order of $DC$ \\ Let $J=\{j_1, \ldots, j_d\}$  be the indices of the words in $V$ with the $d$ largest $DC$s. \\
    Compute document bit-vector signatures \\
    \For{$i=1, \ldots, N$}
    {
    $v(\mathcal{D}_i)  = \{ v_m^{(\mathcal{D}_i)} \}_{m=1}^{d}$\\
    where $v_m^{(\mathcal{D}_i)} =  \mathbbm{1} \{ \tilde{w}_{j_{m}} \in \mathcal{D}_i \cap {j_{m}} \in R_i \}$\\
    and $R_i = \argmin\limits_{R \subset J,|R|=k_2} \sum\limits_{r \in R} DC(\tilde{w}_r)$
    }
}
\caption{\modelname Initial Construction}
\label{alg:construct}
\end{algorithm}

\section{Experiments}\label{sec:exp}
We evaluate on document retrieval and corpus-based set expansion tasks, conducting experiments in two real use-case scenarios on a web-scale collection of documents:
\begin{itemize}
\item \textbf{\textsc{Astronomy}:}
In the first use-case, a research client working on the astronomy field, termed subject matter expert (SME), is interested in collecting documents related to astronomy, capturing a set of - undefined during retrieval - specific entities and phrases $\mathcal{S} = \{s_i\}_{i=1}^{T}$, where $s_i$ is a domain-specific phrase, \eg ``gravitational pull'', ``galactic halos'', etc. The astronomy expert has only provided a few sample documents as relevant examples of what they are looking for. Thus, the query set $\mathcal{Q} = \{Q_i\}_{i=1}^{M}$ is a set of documents (seed corpus). Note that we do not utilize the set $\mathcal{S}$ in our pipeline; set coverage is only used for evaluation purposes. The number of entities and phrases in $\mathcal{S}$ is $496$, while the seed corpus $\mathcal{Q}$ contains $379$ web documents. 
\item \textbf{\textsc{Bottled Water}:}
The second use-case follows the same experimental construction, with the difference that the task is expanding a set of bottled water companies \eg ``Evian'', ``San Pellegrino'', \etc Similarly to the \textbf{\textsc{Astronomy}} experiment, we are provided a few sample documents as relevant examples of what they are looking for. The number of entities is $|\mathcal{S}|=74$ and the seed corpus of sample relevant documents is  $|\mathcal{Q}|=49$. This experiment facilitates evaluation on a limited seed corpus scenario.
\end{itemize}

The document collection is a snapshot (approximately $200$ million documents) of the Common Crawl web archive\footnote{\url{http://commoncrawl.org/}}, which is regularly updated and consists of petabytes of publicly hosted webpages. As far as our evaluation metrics, we check the coverage of the phrase set $\mathcal{S}$, \ie how many of the set phrases are found in the retrieved top-$k$ most relevant documents, as well as the total query execution time from each method. 

We compare against traditional vector representation methods and pre-trained language models that produce dense contextualized representations:
\begin{itemize}
    \item \textbf{Random} selection of a subset of $k$ documents.
    \item \textbf{TF-IDF} traditional bag-of-words (TF-IDF) model.
    \item \textbf{Hash} vector representations that utilize the hashing trick to reduce the dimensionality \cite{weinberger2009feature}.
    \item \textbf{RoBERTa} pre-trained language model, in particular, a distilled version of the RoBERTa model \cite{Sanh2019DistilBERTAD}, trained on the MSMARCO Passage Ranking dataset \cite{bajaj2016ms}, \ie this model is specifically designed and pre-trained for semantic search.
    \item \textbf{\modelname}, \ie our proposed method for designing document signatures.
    \item  \textbf{Query} is a lexicon-based search where ``in hindsight'' we use a sample of the set $S$ as queries and retrieve documents that match the terms. This method follows a concept expansion query setting, \ie given a couple of seed entities belonging to the concept, retrieve more entities that belong to the same concept. Albeit the query here is formulated differently than the rest of the methods, this comparison is particularly useful to showcase the limitations of set expansion methods that lack context.
\end{itemize}
\begin{algorithm}[t!]
\DontPrintSemicolon
\textbf{Input:}{\\Corpus $\mathcal{C}$, vocabulary $V$, document frequency counts $DC$ \\
$v(\mathcal{D}_i)  = \{ v_m^{(\mathcal{D}_i)} \}_{m=1}^{d}, \forall \mathcal{D}_i \in \mathcal{C}$: signatures \\
$k_1, k_2$: document thresholds \\
$d$: signature dimension \\
$\bar{\mathcal{D}} = \{\bar{w}_1, \ldots, \bar{w}_{\bar{T}} \}$:  new document\\}
\Begin{
    Update vocabulary\\ $\hat{V} = V \cup \{\bar{w}_1, \ldots, \bar{w}_{\bar{T}} \} = \{\hat{w}_1, \ldots, \hat{w}_{|\hat{V}|}\}$\\
    Update document counts $DC$ \\
    \For{each word $\hat{w}_m \in \bar{\mathcal{D}}$}
    {
    $DC(\hat{w}_m) =  DC(\hat{w}_m) + 1 \hide{\mathbbm{1} \{\hat{w}_m \in  \hat{\mathcal{D}}\}}$ \\
    where $DC(\hat{w}_m) = 0 $ if $\hat{w}_m \notin V$
    }
    Update signature dimension $\hat{d} = |\hat{V}| - \sum\limits_{m=1}^{\hat{V}} \mathbbm{1} \{DC(\hat{w}_m) < k_1\}$\\
    Update signature vectors accordingly.
}
\caption{\modelname Stream Updates}
\label{alg:update}
\end{algorithm}

\subsection{Computing Infrastructure}
We make use of a $64$-bit Ubuntu $16.04.6$ machine with Intel Xeon Gold CPU @$2.10$GHz ($44$ cores/$88$ threads), $3$TB RAM and a Tesla V100 Nvidia GPU. 
For RoBERTa, our model is based on the HuggingFace Transformers and Sentence-BERT implementations \cite{Wolf2019HuggingFacesTS,reimers-2019-sentence-bert}.  Note that encoding each document with pre-trained language models is a fairly expensive procedure\footnote{https://github.com/hanxiao/bert-as-service}. We store the document vectors in advance and do not count the encoding time on our experiments, but we observe that our proposed method performs much better in terms of encoding latency.

\subsection{Hyper-parameter details}
In our experiments, we set $k=500,000$, \ie retrieving $500,000$ relevant documents. This is a sufficiently large set to perform analysis in most downstream systems.
Moreover, the document thresholds are set to $k_1 = 1000$ and $k_2 =100$, based on the subject matter expert's intuition and suggestions. As pre-processing step, all methods require to build and store document signatures, \ie vector representations for each document in the collection. For fair comparison, we set the same time limits for vector construction and stop the process when the time is exhausted. 
The dimensionality of all document vector representation methods is set to $d=100$. For RoBERTa, we perform PCA to reduce the dimensionality of the pre-trained vectors. For the lexicon-based search, we use $10\%$ of the set phrases as queries, which equals 50 keyword phrases for \textsc{\textbf{Astronomy}} and 7 keyword phrases for \textsc{\textbf{Bottled Water}}. For both the non-deterministic strategies, \ie lexicon-based search and random search, we report the average query execution time and coverage across 3 independent trials. All baselines are similarly parallelized.\\

\noindent \textbf{RAM/Disk Comparison:} In terms of memory and space footprint, \modelname is very efficient, as each document signature is 400 bytes (100 non-zero entries, 4 bytes each). In comparison, each document consumes approximately 1,250 bytes for Hash, 1000 bytes for TF-IDF, and 528 bytes for 
RoBERTa. With this reduction in size of compared vectors we achieve the obvious improvement in performance and storage needs. With the terms in the sparse SAUCE vector pre-sorted, the comparison becomes a very fast sequential memory (and disk) access to perform a classic MERGE AND operation.

\begin{figure}[t!]
\centering
\subfigure[\textsc{\textbf{Bottled Water}}]{
\includegraphics[width=0.486\columnwidth]{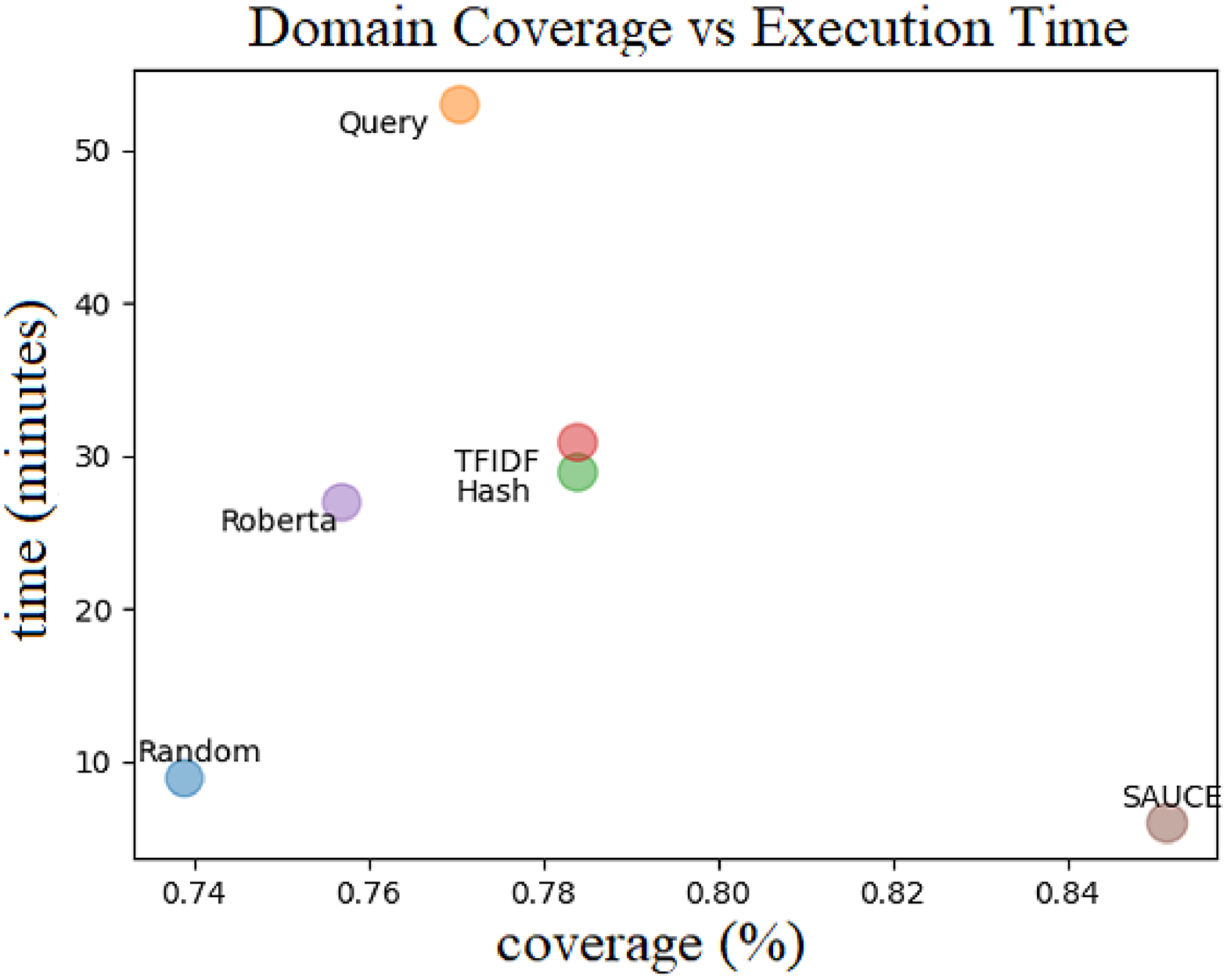}}
\subfigure[\textsc{\textbf{Astronomy}}]{
\includegraphics[width=0.486\columnwidth]{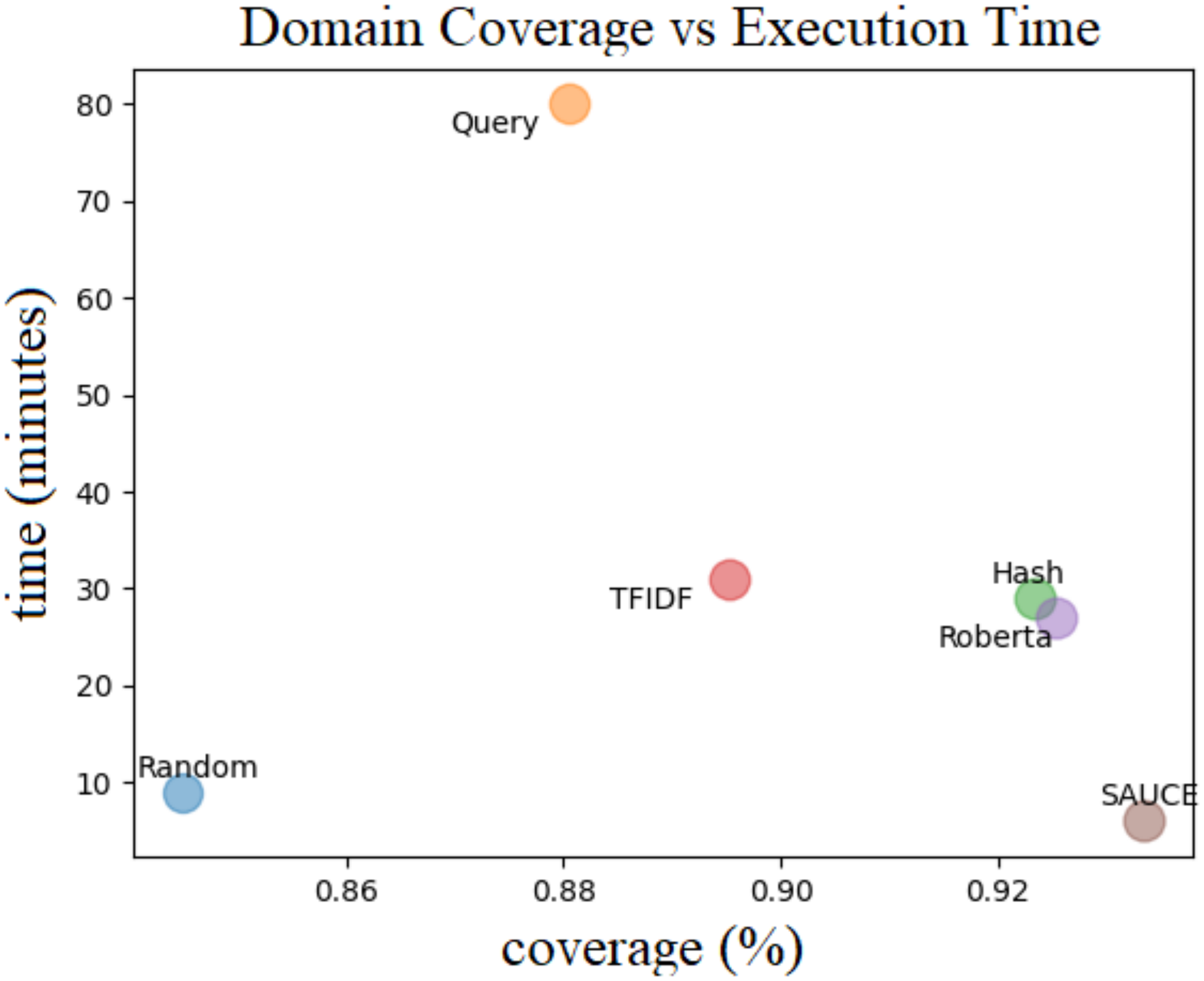}}
\caption{Trade-off between coverage vs. query execution time for \textsc{\textbf{Bottled Water}} (left) and \textsc{\textbf{Astronomy}} (right). Best viewed in color. Compared to baselines, \modelname produces better domain coverage in a fraction of the execution time.} 
\label{fig:results}
\end{figure}

\begin{figure*}[t!]
\centering
\subfigure[\modelname]{
\includegraphics[width=0.65\columnwidth]{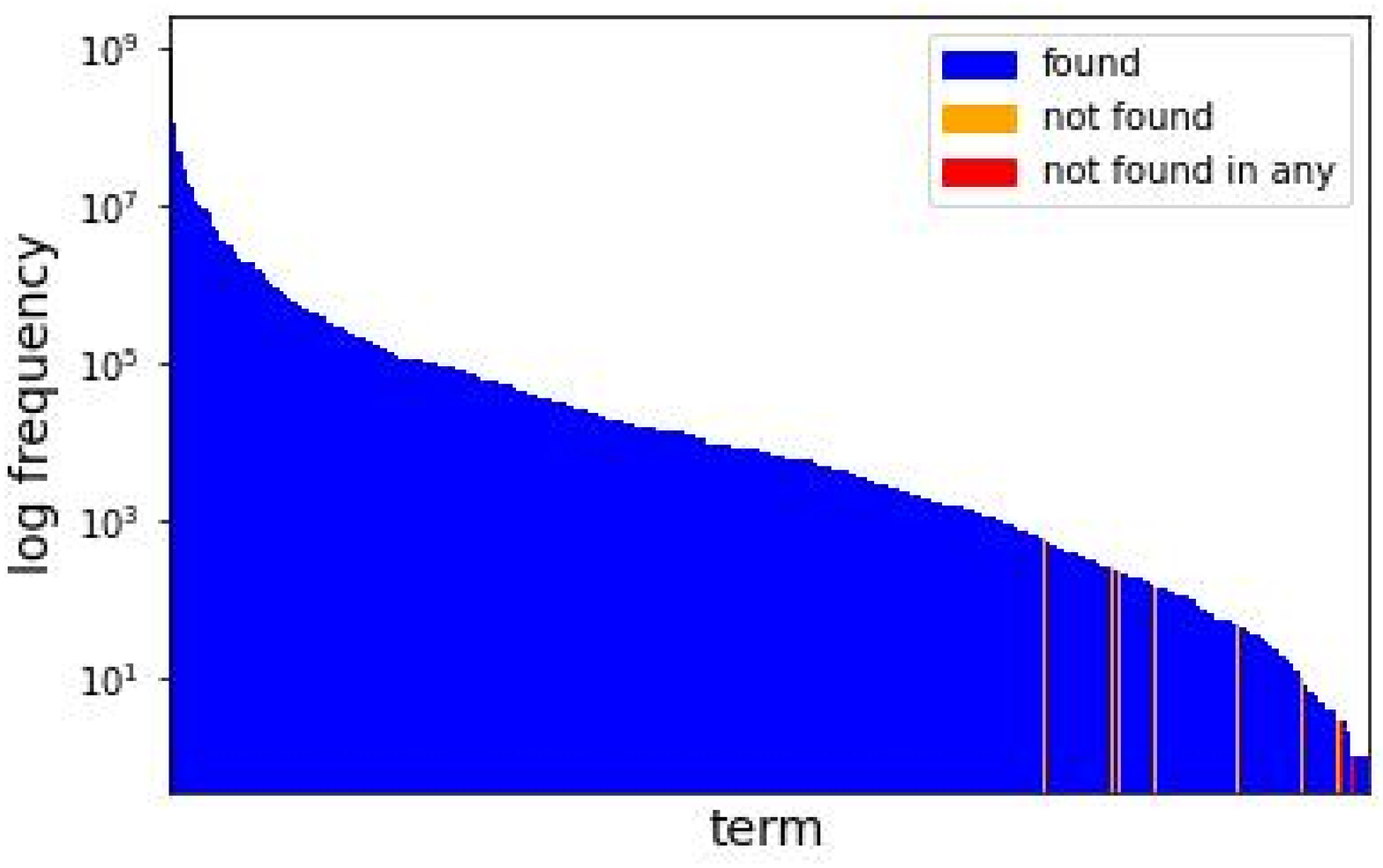}}
\subfigure[RoBERTa]{
\includegraphics[width=0.65\columnwidth]{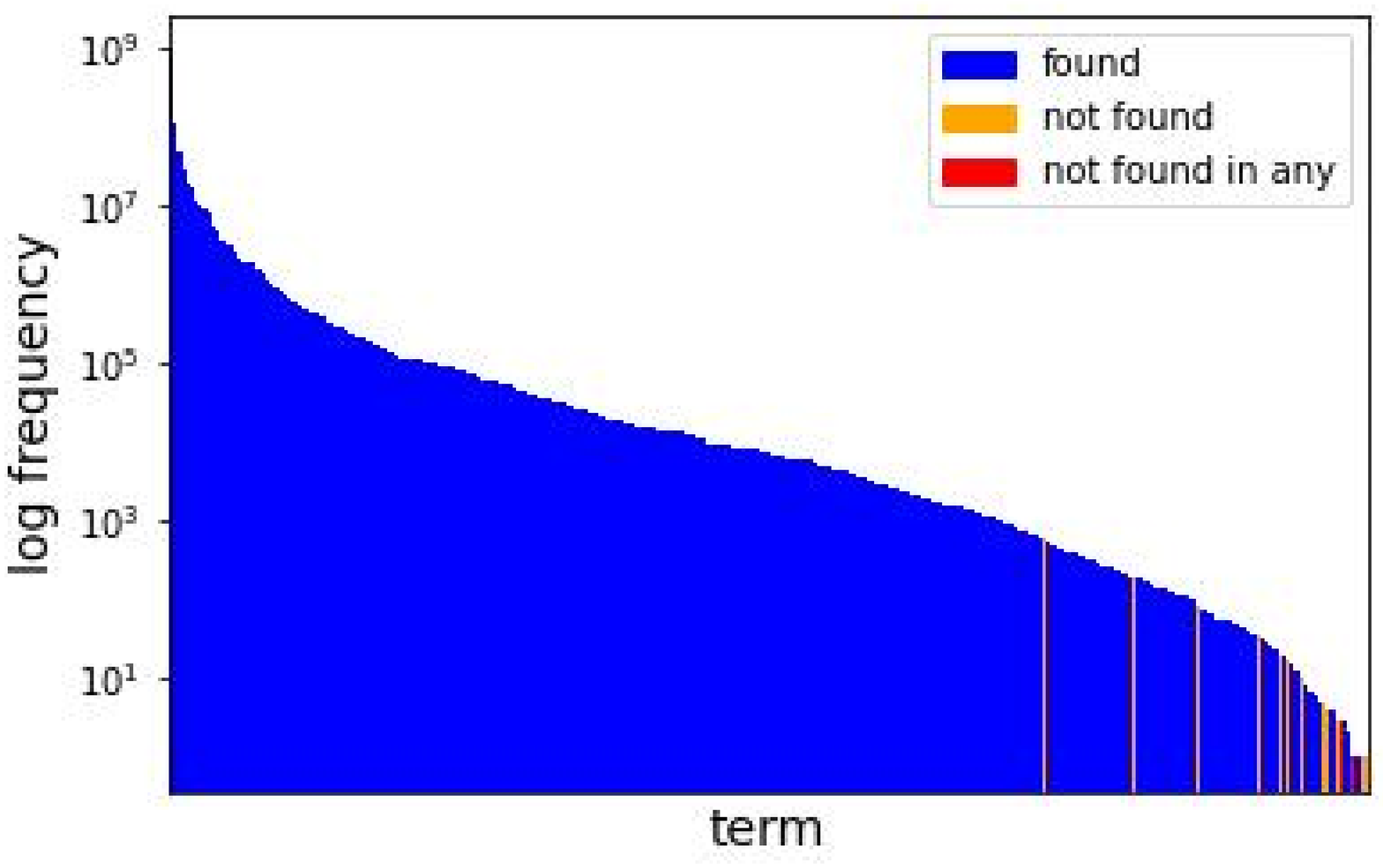}}
\subfigure[Hash]{
\includegraphics[width=0.65\columnwidth]{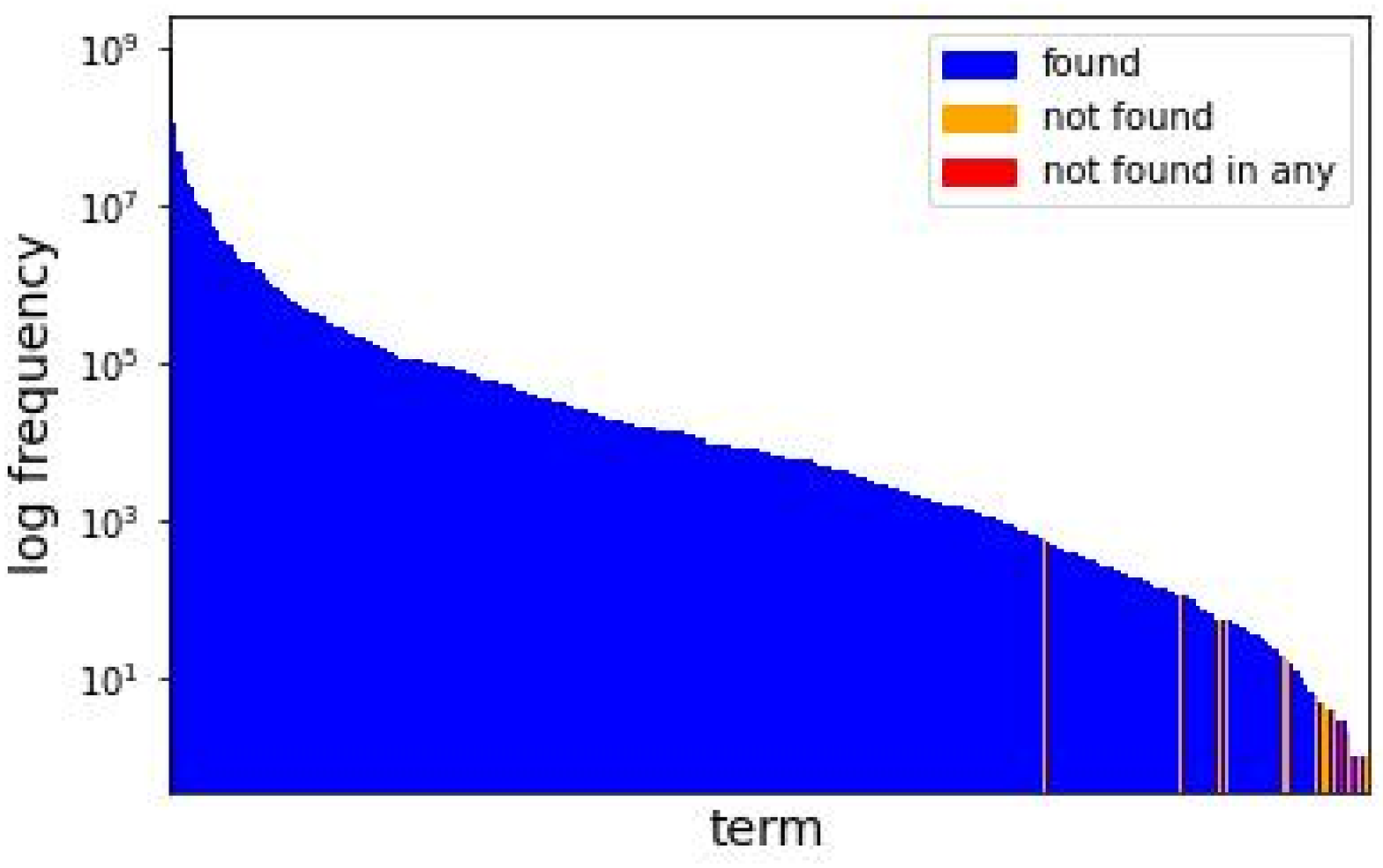}}
\subfigure[TF-IDF]{
\includegraphics[width=0.65\columnwidth]{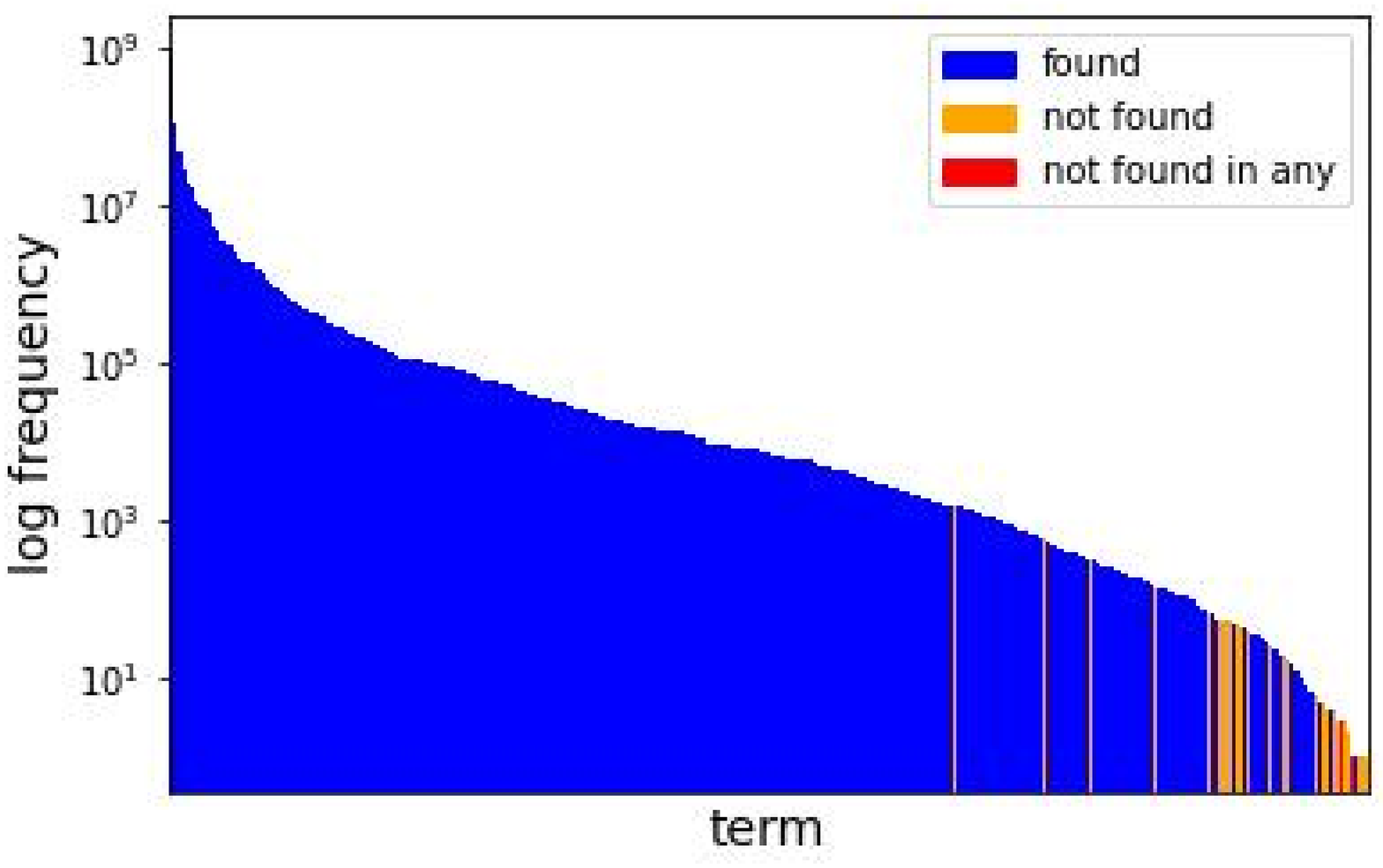}}
\subfigure[Query]{
\includegraphics[width=0.65\columnwidth]{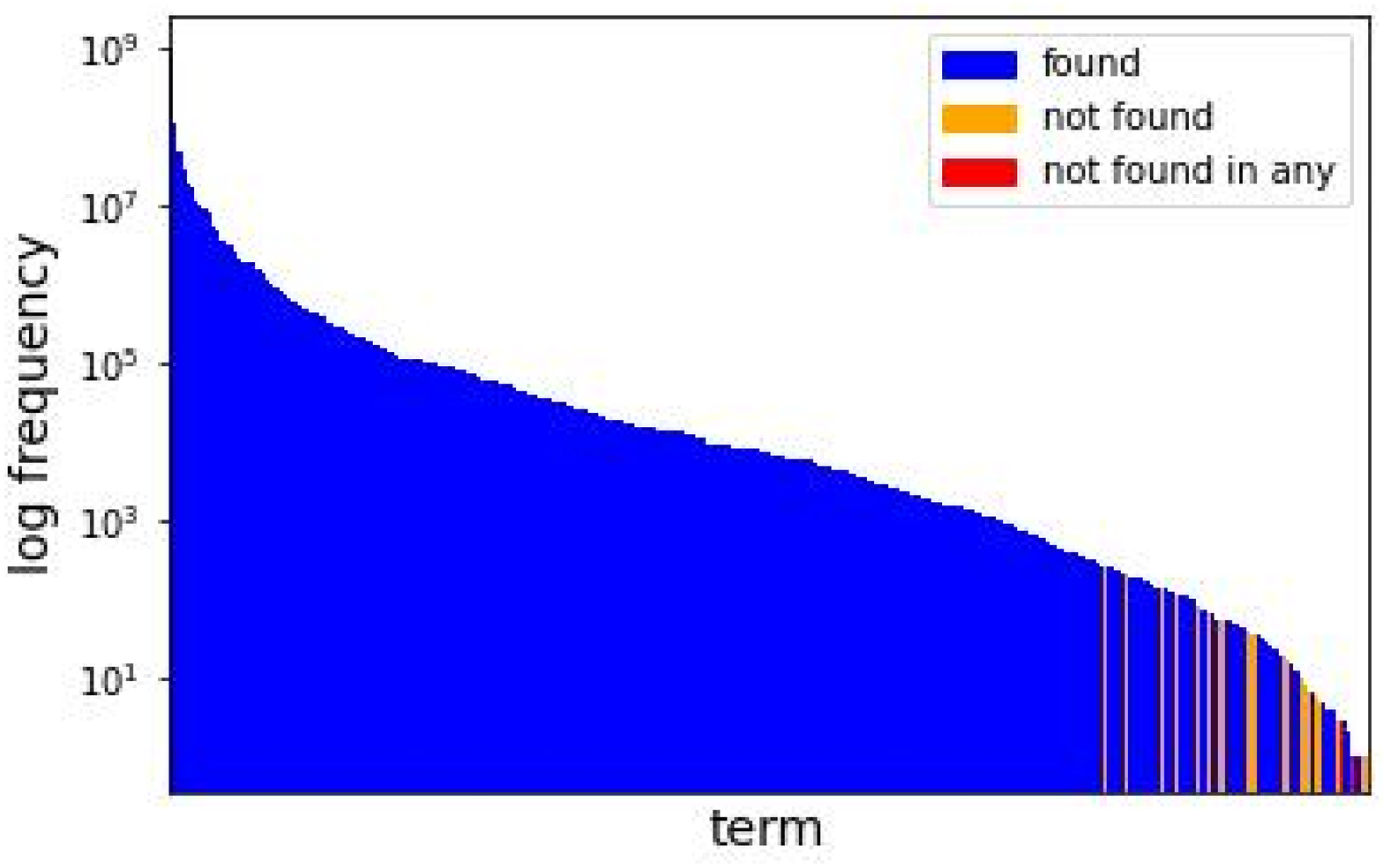}}
\subfigure[Random]{
\includegraphics[width=0.65\columnwidth]{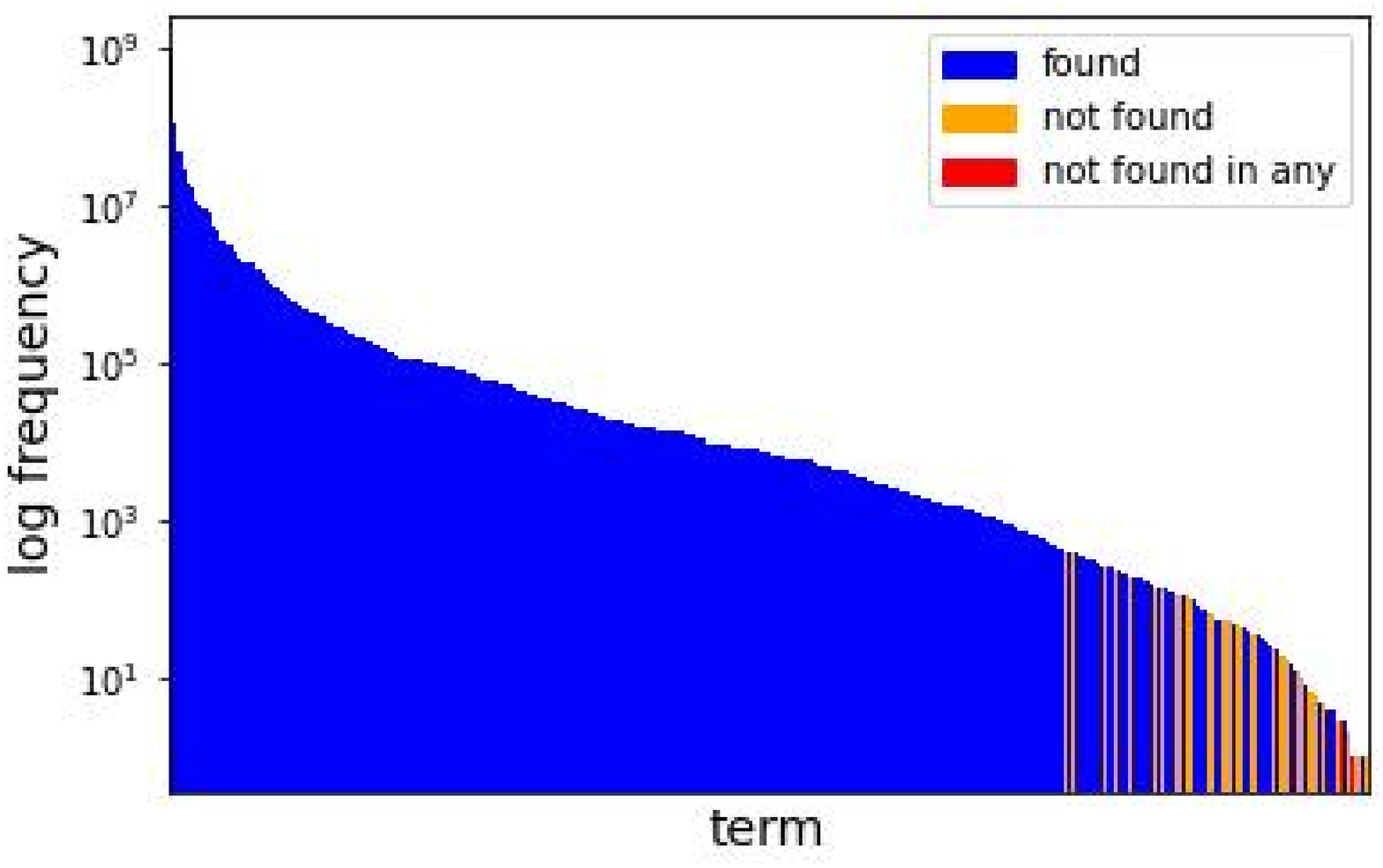}}
\centering
\caption{Comparison of methods with frequency histogram log-plots  in the \textsc{\textbf{Astronomy}} domain. Best viewed in color. Orange indicates a term not found by retrieving documents with the respective method, blue indicates terms found, and red outlines terms not found with any of the compared methods. Length of each bar indicates term frequency, for example Hash missed a few rare terms while \modelname has lower coverage of the long tail.
} 
\label{fig:astro_histograms}
\end{figure*}

\subsection{Web-scale Concept Expansion Results}
Given a few seed entities as input, web-scale concept expansion deals with the task of finding the complete set of entities that belong to the same conceptual class, by analyzing the retrieved documents when querying either with the seed entities or with a seed corpus of relevant documents.

In Figure \ref{fig:results}, we plot the query execution time versus the domain lexical coverage, as captured by the number of phrases from set $\mathcal{S}$ found in the top-$k$ retrieved documents. In general, we find that \textbf{Query} is the least efficient method in terms of query execution time and has low overall lexical coverage ($30$ mins and $77.8\%$ coverage, and $80$ minutes with $88.04\%$ coverage, for \textsc{\textbf{Astronomy}} and \textsc{\textbf{Bottled Water}}, respectively). 

On the other hand, for \textsc{\textbf{Astronomy}}, the \textbf{RoBERTa} dense representations can retrieve a smaller set of documents that are highly specialized within the domain at-hand ($92.54\%$ coverage and $27$ minutes execution time). However, \textbf{RoBERTa} seems to produce suboptimal results for \textsc{\textbf{Bottled Water}}, with \textbf{Query} surpassing \textbf{RoBERTa}. We find that the more distinct the concept, with more named entities, \eg ``San Luis'', ``Propel Fitness Water'' ``Spa'', ``Voss'', \etc, the better the \textbf{Query} performance is.

Our method, \textbf{\modelname}, surpasses all baselines by a large margin, with $93.35\%$ coverage and $6$ minutes execution time in total. In other words, \textbf{\modelname} reduces the query time by $78\%$ from the next best strategy (w.r.t. lexical coverage), while simultaneously increasing performance up to $6.8\%$. In the future, we hope to deploy the system to other real-world scenarios and investigate the performance in other domains, working with experts in a diverse set of research fields.

\begin{figure*}[t!]
\centering
\subfigure[\modelname]{
\includegraphics[width=0.65\columnwidth]{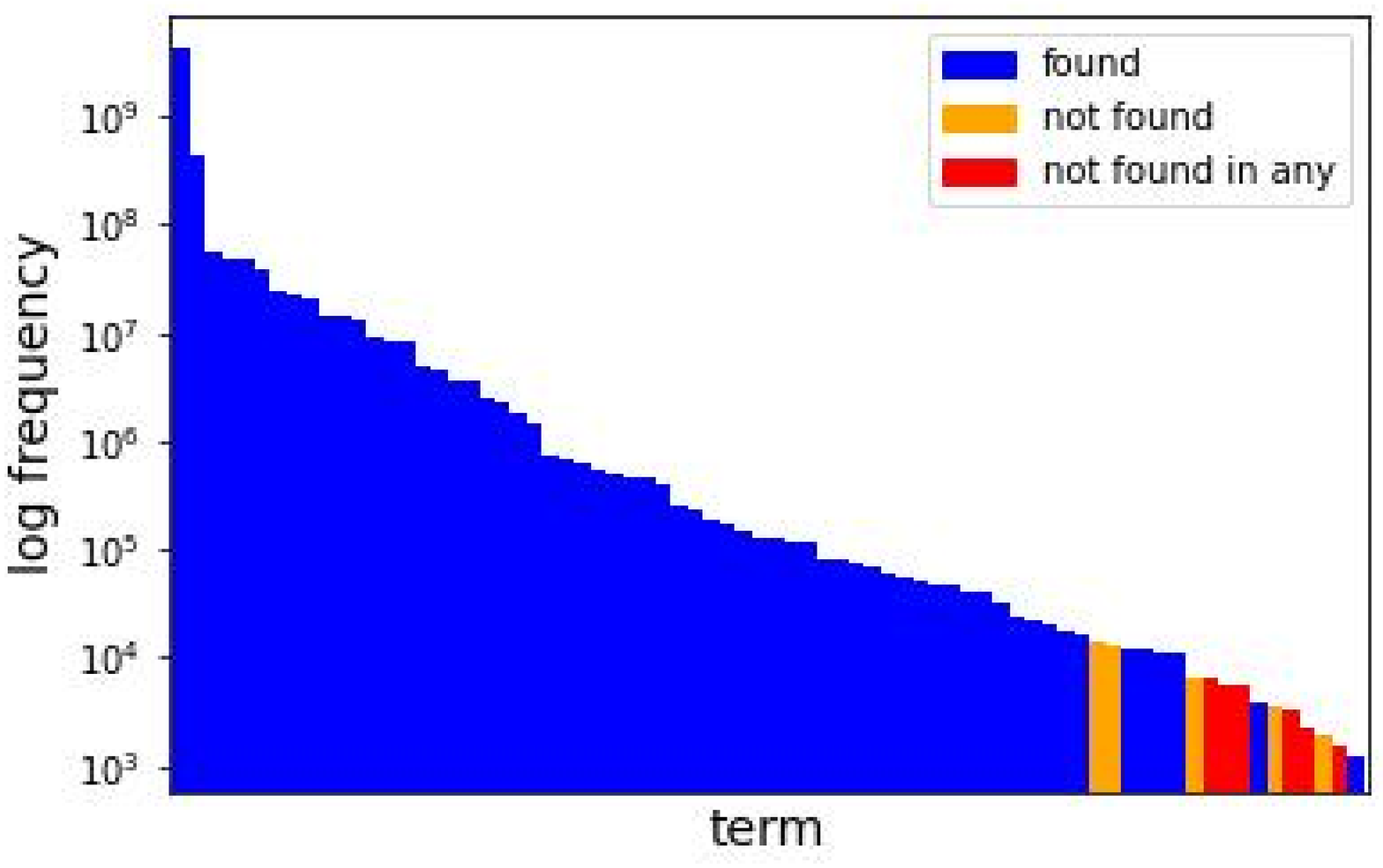}}
\subfigure[RoBERTa]{
\includegraphics[width=0.65\columnwidth]{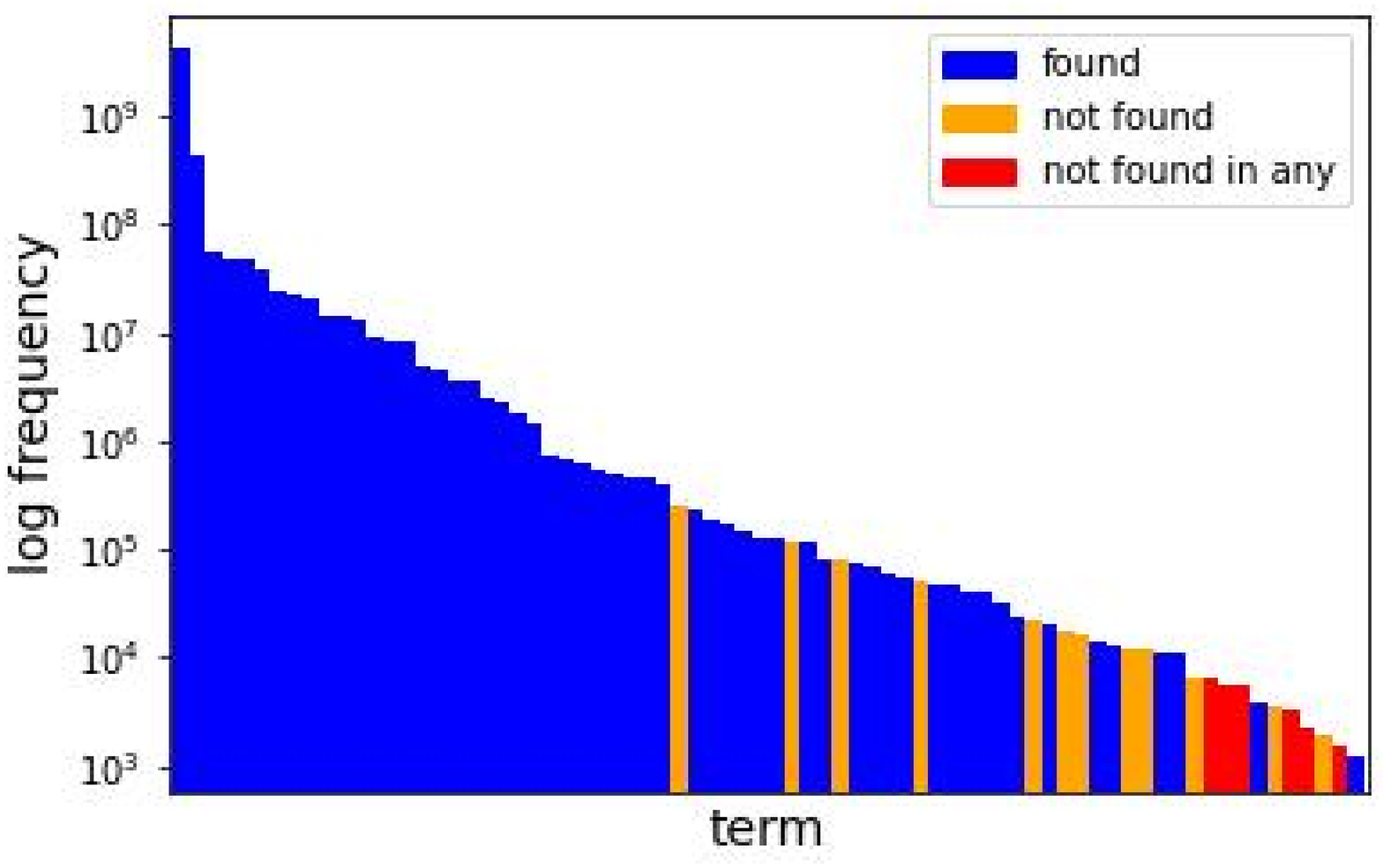}}
\subfigure[Hash]{
\includegraphics[width=0.65\columnwidth]{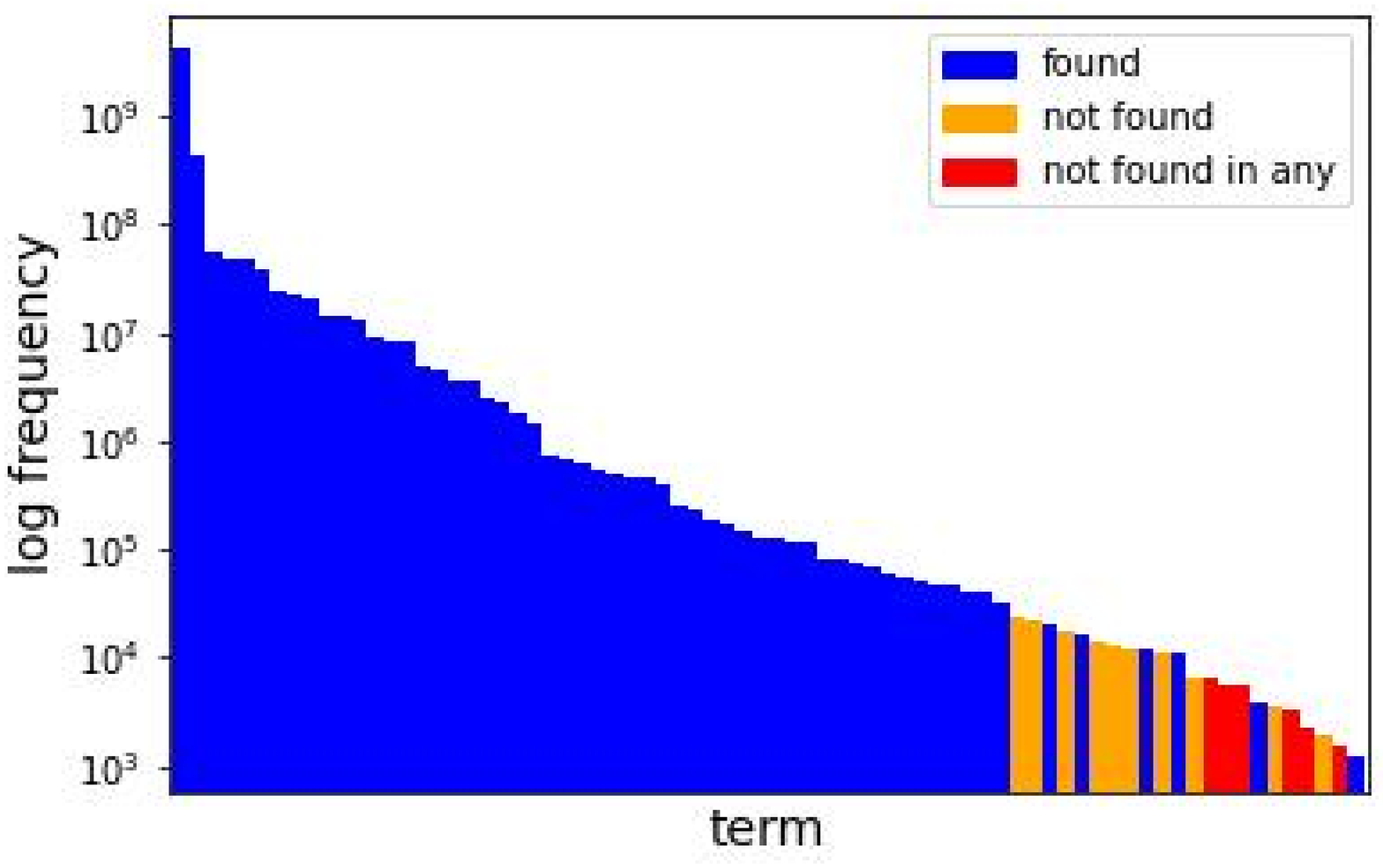}}
\subfigure[TF-IDF]{
\includegraphics[width=0.65\columnwidth]{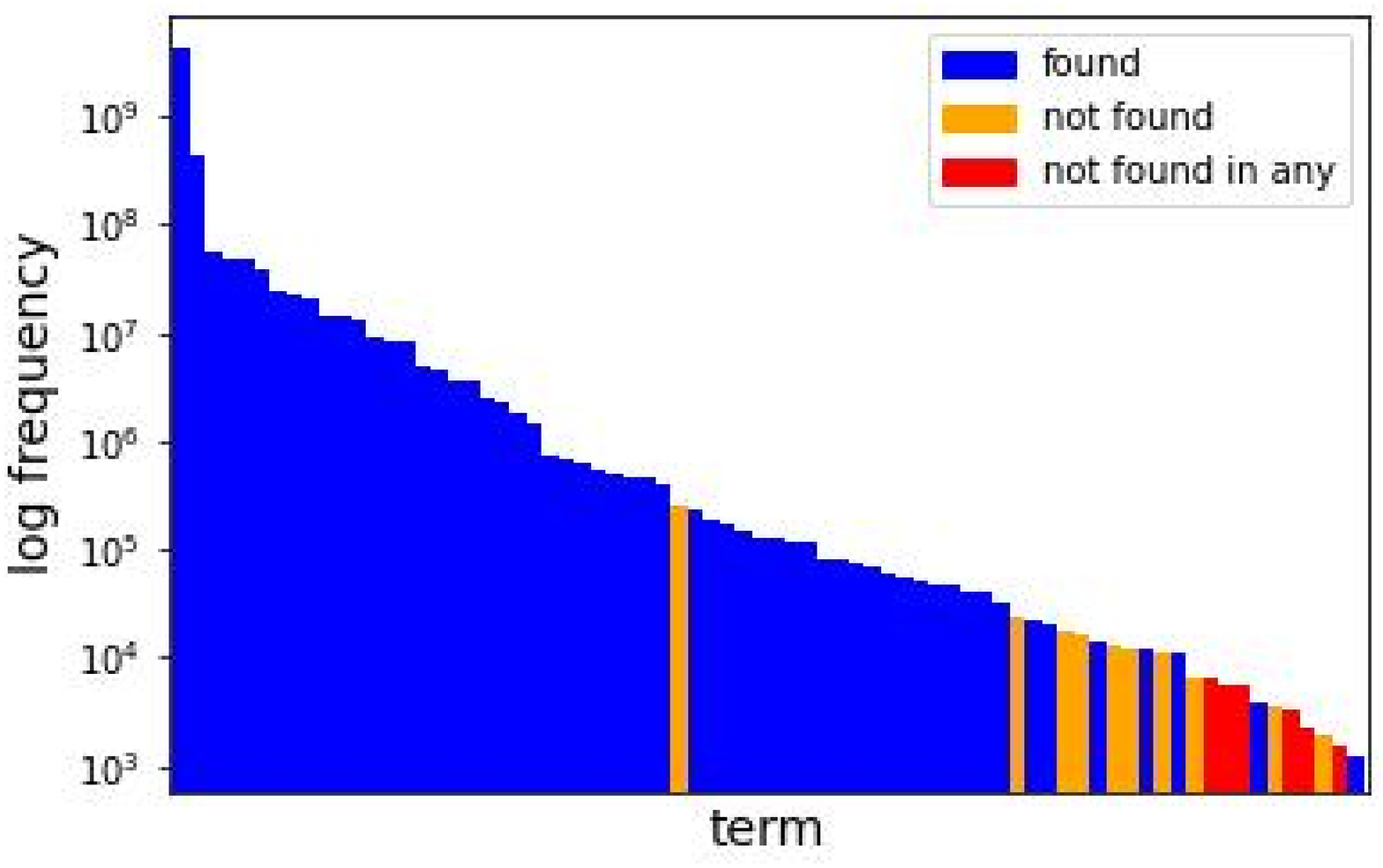}}
\subfigure[Query]{
\includegraphics[width=0.65\columnwidth]{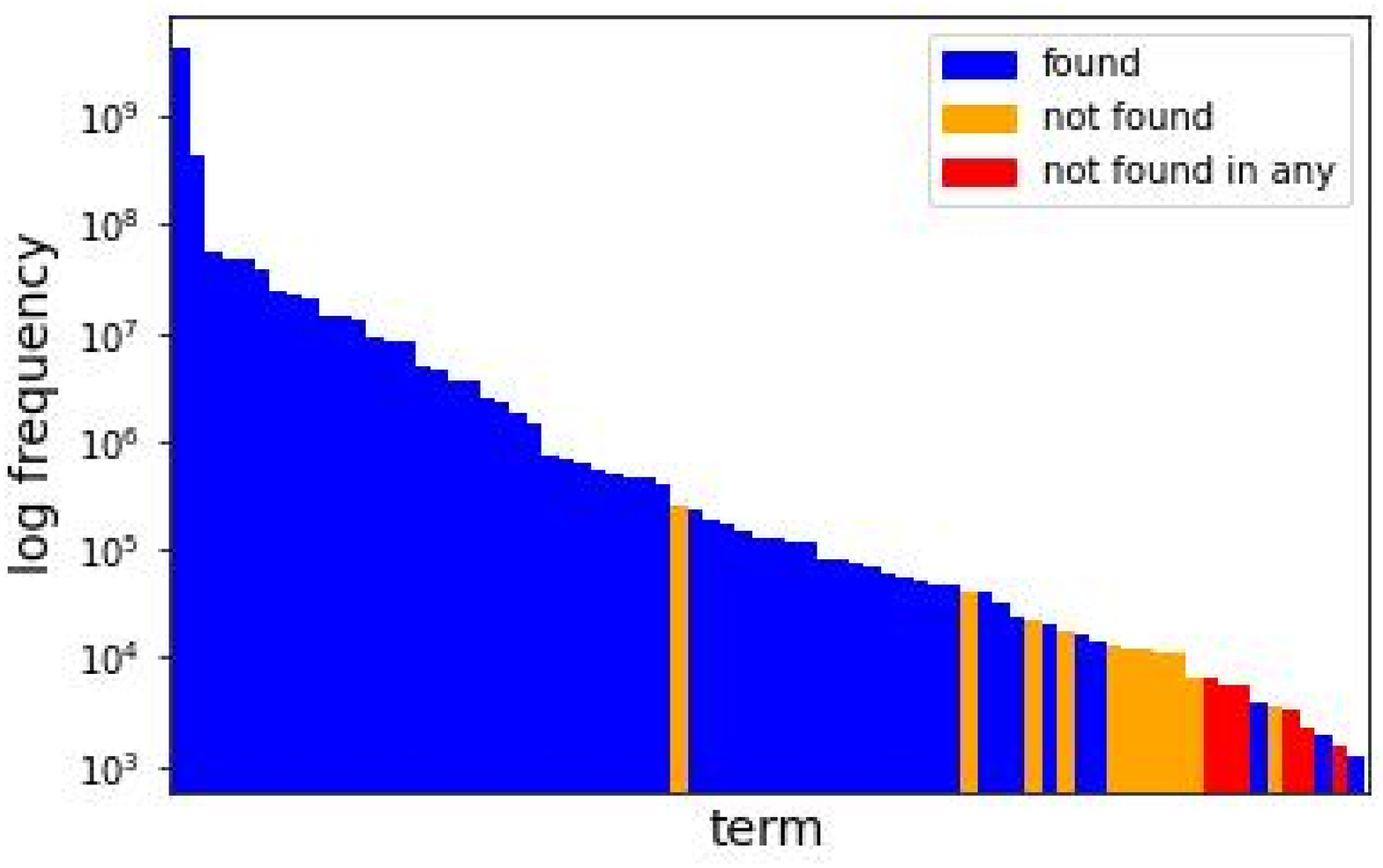}}
\subfigure[Random]{
\includegraphics[width=0.65\columnwidth]{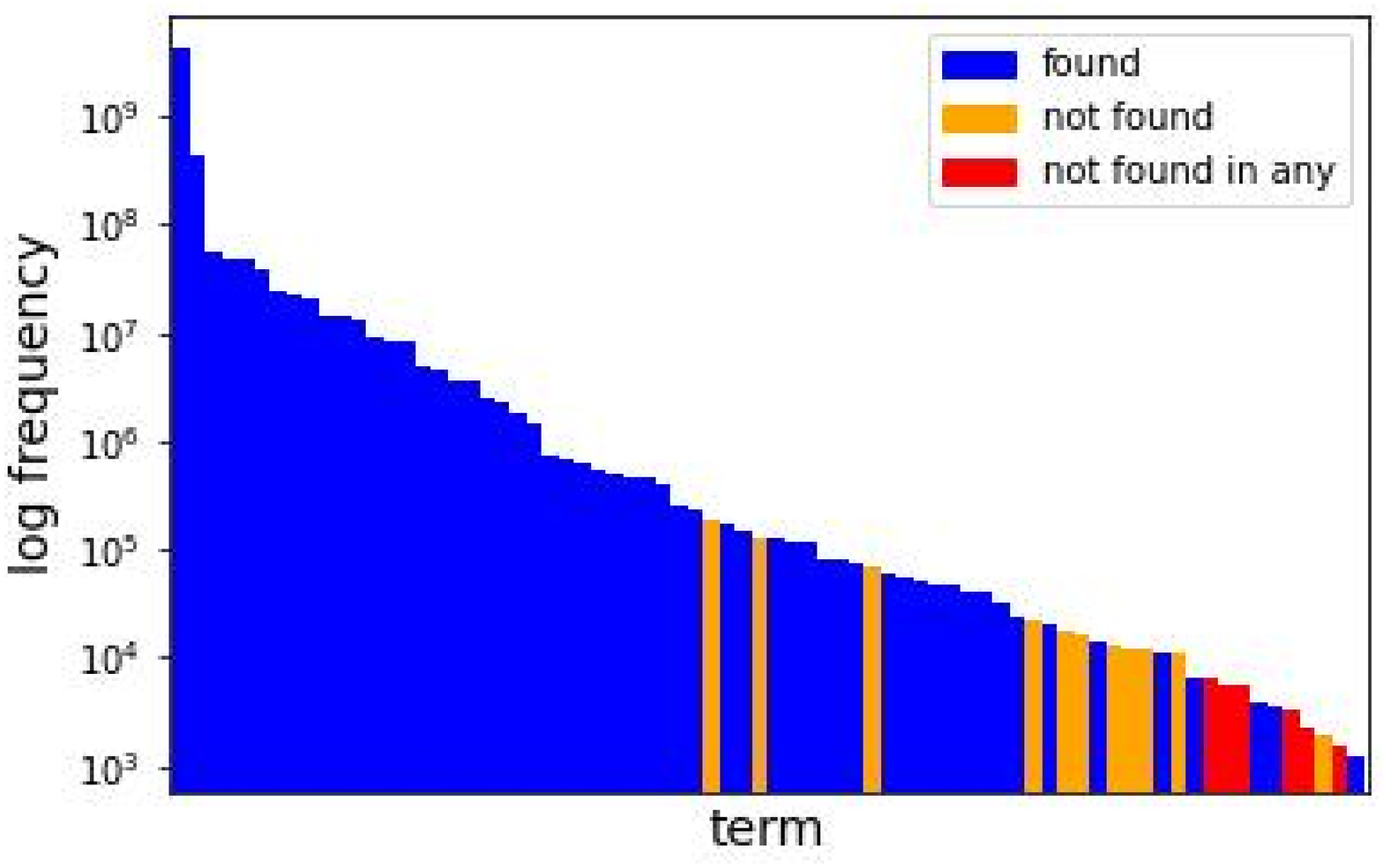}}
\centering
\caption{Comparison of methods with frequency histogram log-plots  in the \textsc{\textbf{Bottled Water}} domain. Best viewed in color. Orange indicates a term not found by retrieving documents with the respective method, blue indicates terms found, and red outlines terms not found with any of the compared methods. Length of each bar indicates term frequency.
\modelname missed fewer rare terms, better capturing the long tail.} 
\label{fig:water_histograms}
\end{figure*}
\begin{table*}[t]
\resizebox{\textwidth}{!}{
\begin{tabular}{c|lr|lr|lr|lr|lr}
\toprule
\multirow{2}{*}{\textbf{Dataset}} & \multicolumn{2}{c}{\textbf{SAUCE}} & \multicolumn{2}{c}{\textbf{Hash}} & \multicolumn{2}{c}{\textbf{TF-IDF}} & \multicolumn{2}{c}{\textbf{RoBERTa}} & \multicolumn{2}{c}{\textbf{Query}} \\  \cline{2-11} 
\multirow{9}{*}{\textsc{\textbf{Astronomy}}} & \textbf{Term} & \textbf{Freq.} & \textbf{Term} & \textbf{Freq.} & \textbf{Term} & \textbf{Freq.} & \textbf{Term} & \textbf{Freq.}  & \textbf{Term} & \textbf{Freq.}  \\ \hline

 & lunt filter         & 579   & lunt filter         & 579   & optical spectroscopy    & 1533  & lunt filter           & 579   & subgiant            & 265   \\
& iotian              & 250   & galactic halos      & 117   & lunt filter             & 579   & skathi                & 193   & exoplanets orbiting & 208   \\
& ci tau              & 227   & circumplanetary     & 56    & active galactic nucleus & 312   & a journey to the moon & 81    & ultracool dwarf     & 135   \\
& gamma emission      & 143   & young massive stars & 53    & gegenschein             & 180   & apastron              & 61    & superflares         & 134   \\ \cline{2-11}
& f - stop            & 45    & astronomik filters  & 2     & astronomik filters      & 2     & jarnsaxa              & 4     & background cmb      & 3     \\
& sky , which         & 10    & mundilfari          & 1     & mundilfari              & 1     & background cmb        & 3     & mundilfari          & 1     \\
& space exploration & 10    & kiviuq              & 1     & kiviuq                  & 1     & mundilfari            & 1     & ngc691              & 1     \\
& background cmb      & 3     & ngc691              & 1     & ngc691                  & 1     & thrymr                & 1     & thrymr              & 1 \\  
\cline{1-11} \cline{1-11}

\multirow{9}{*}{\textsc{\textbf{Bottled Water}}} & deep river rock & 248000 & propel fitness water & 23300 & deep river rock & 248000 & deep river rock & 248000 & deep river rock & 248000 \\
& agua vida & 121000 & ethos water & 22000 & propel fitness water & 23300 & agua vida & 121000 & harrogate spa water & 39900 \\
& dejà blue & 78600 & gerolsteiner brunnen & 18100 & gerolsteiner brunnen & 18100 & dejà blue & 78600 & ethos water & 22000 \\
& trump ice & 52000 & neviot & 14100 & souroti & 16700 & trump ice & 52000 & gerolsteiner brunnen & 18100 \\ 
\cline{2-11}
& rhonsprudel & 13200 & fruit2o & 11000 & fruit2o & 11000 & bling h2o & 11700 & fruit2o & 11000 \\
& w8 water & 6730 & w8 water & 6730 & w8 water & 6730 & w8 water & 6730 & iceland pure spring water & 10900 \\
& hiram codd & 3480 & hiram codd & 3480 & hiram codd & 3480 & hiram codd & 3480 & w8 water & 6730 \\
& knjaz miloš ad & 2000 & knjaz miloš ad & 2000 & knjaz miloš ad & 2000 & knjaz miloš ad & 2000 & hiram codd & 3480 \\
 
\bottomrule
\end{tabular}%
}
 \caption{Most-common and least-common terms not found in retrieved documents.}
  \label{table:missed}
\end{table*}

An important factor in successfully capturing a topic concept is the domain coverage, and corpus collection methods should properly capture the long tail of domain-specific rare terms. To showcase the advantage of our method, we present a comparison with term frequency histogram log-plots, where for each term a bar plot (the length of the bar plot) indicates the term frequency. In Figures \ref{fig:astro_histograms} and \ref{fig:water_histograms}, the color of each bar plot indicates whether a term was found (blue) or not (orange) in the top-$k$ ranked documents, based on each respective method. We also highlight terms not found by any of the compared methods with red color. \textbf{Random},  \textbf{Query} and \textbf{TF-IDF} appear more orange in the lower tail of the distribution, while \textbf{RoBERTa} performs the best among the baselines. Compared to all methods, \textbf{\modelname} seems to be missing the least number of rare terms, better capturing the long tail.

\begin{figure}[t!]
\centering
\subfigure[\textsc{\textbf{Astronomy}}]{
\includegraphics[width=0.48\columnwidth]{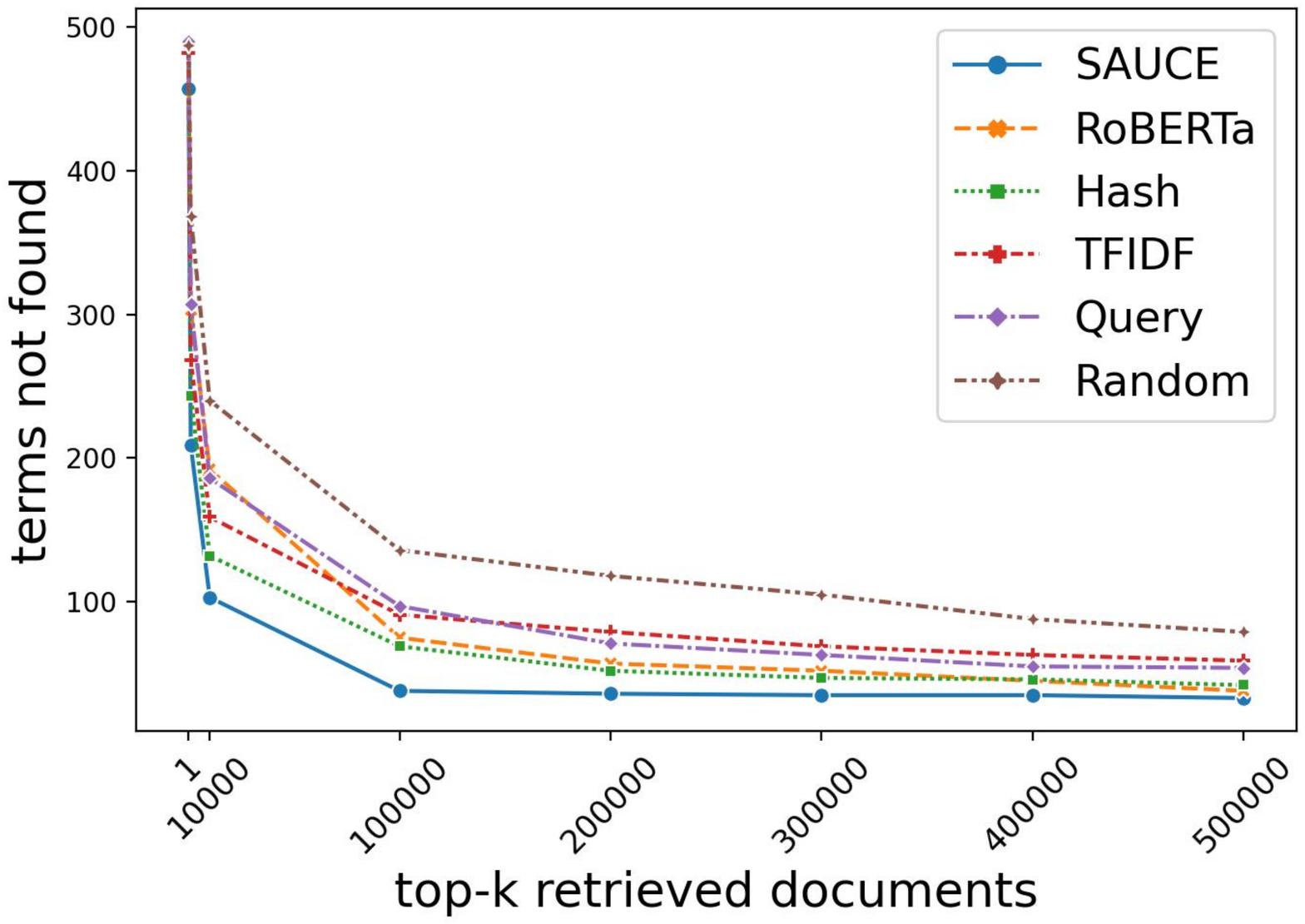}}
\subfigure[\textsc{\textbf{Bottled Water}}]{
\includegraphics[width=0.48\columnwidth]{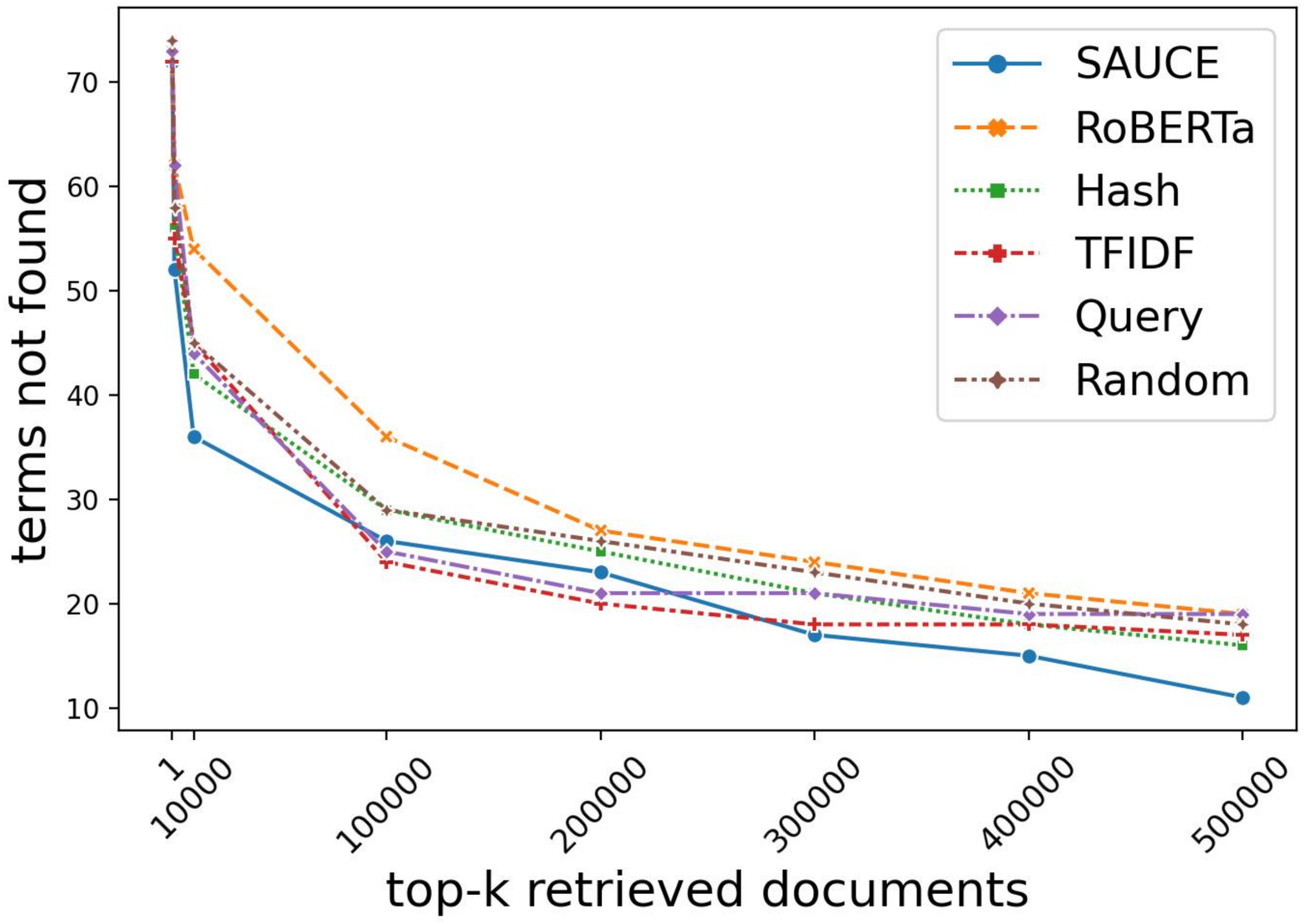}}
\centering
\caption{Ablation study on the number of lexicon entries not found as the as the of retrieved documents (top-$k$) grows for both the \textsc{\textbf{Astronomy}} and \textsc{\textbf{Bottled Water}} domains. 
} 
\label{fig:ablation}
\end{figure}

Finally, in Table \ref{table:missed} we present document counts for the most common and least common terms missed by each method, and in Figure \ref{fig:ablation} we present an ablation study on the terms missed as the number of documents retrieved (top-$k$) grows. Notably, the recall of \textbf{\modelname} (blue line) gets better as more documents are retrieved, with increasing lexical domain coverage, showing that \textbf{\modelname} can better capture the long tail in corpus expansion tasks.

\subsection{Document Retrieval Results}
\begin{figure}[t!]
\centering
\includegraphics[width=0.9\columnwidth]{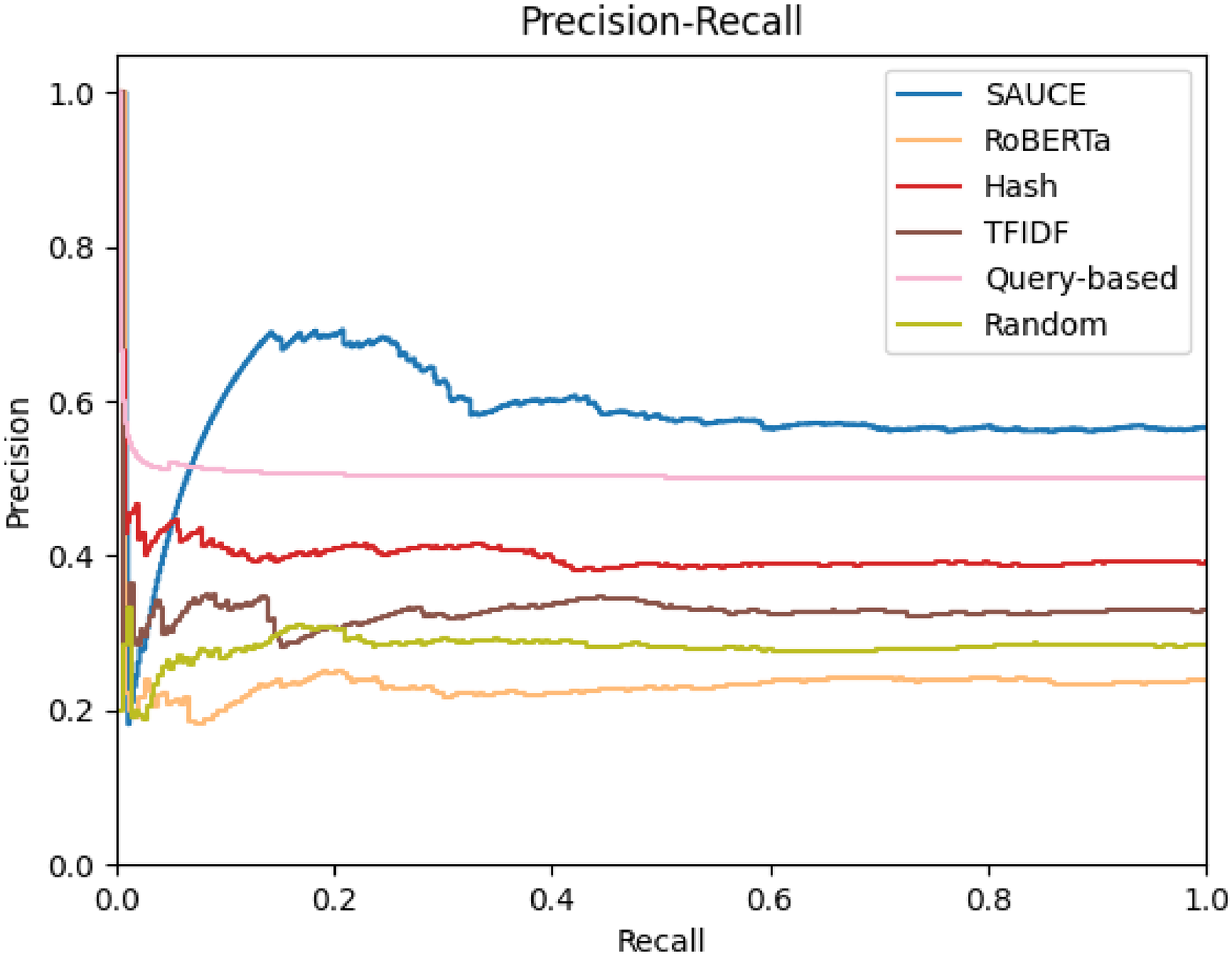}
\centering
\caption{Precision-Recall Curves for all compared methods, on the \textsc{\textbf{Bottled Water}} document retrieval task. Best viewed in color.
} 
\label{fig:prcurve}
\end{figure}

We also evaluate the relevance of the retrieved documents. For the \textbf{\textsc{Bottled Water}} domain, we collect the top-$1000$ documents and ask a subject matter expert (SME)  to provide a binary judgment (relevant or non-relevant) for each document in the  top-$1000$ retrieved results. In Table \ref{tab:retrieval_results}, we report 
Normalized Discounted Cumulative Gain (nDCG), Recall, and Mean Average Precision (MAP). We additionally present precision-recall curves (Figure \ref{fig:prcurve}). \textbf{\modelname} achieves higher MAP, Recall and nDCG scores than the rest of the methods, with \textbf{Query} being the next best strategy. Albeit a small sample of relevance judgments, overall our results validate that sparse truncated bit-vector document signatures are an efficient representation for fast corpus expansion, with improved long tail lexical coverage.

\section{System Deployment and Use}\label{sec:deploy}
\modelname has been deployed since November 2020, with a patent filed and pending~\cite{gruhl2020}. \modelname is typically used by IBM consultants. The end customer then is provided with the focused collection to be used in their unsupervised learning task. We have used \modelname for a variety of projects with customers expanding small to medium seed corpora. Additional example use cases include ``cell phone issues'' for call centers, ``DevOps'' for software engineering, and ``Pizza'' (the later test set was for a client who didn't want to reveal their actual interest).

Over the last few months, customers have expressed positive feedback for our proposed corpus expansion method. The automatic expanded large-scale corpora have been used to train several machine learning (ML) systems across a variety of use-cases, and in all cases, the ML model performance has been substantially improved over training on just the client-supplied seed corpus.

The biggest difference is the substantially more complete coverage of long tail terms; these include both infrequently used terms as well as terms that are emerging and gaining in popularity. The later set is very critical to track for many of our clients. For example, IBM Garage\footnote{\url{https://www.ibm.com/garage}}, \ie the IBM prototype systems team, integrated a Human-in-the-Loop User Interface to create a social media monitoring application for a business client.

\begin{table}[t!]
\caption{Comparison of \modelname with baselines on the document retrieval task for  \textsc{\textbf{Bottled Water}}. Evaluation based with NDCG, Recall and scores.}
 \centering
\begin{tabular}{c|ccc}
 \toprule
 \multirow{2}{*}{\textbf{Method}} & \multicolumn{3}{c}{\textsc{\textbf{Bottled Water}}} \\ \cline{2-4}
 &  \textbf{{nDCG}}  & \textbf{{Recall}} & \textbf{{MAP}} \\
 \midrule \midrule
 \textbf{\modelname} & \textbf{0.895} & \textbf{0.566} & \textbf{0.584} \\
 \textbf{RoBERTa}   & 0.724 & 0.237 & 0.234\\
 \textbf{TF-IDF}  & 0.786 & 0.329 & 0.329\\
 \textbf{Hash}  & 0.828 & 0.391 & 0.400\\
 \textbf{Random}  & 0.746 & 0.285 & 0.282 \\
 \textbf{Query}  & 0.871 & 0.501 & 0.506\\
\bottomrule
 \end{tabular}
\label{tab:retrieval_results}
\end{table}

\section{Conclusion and Future Work}\label{sec:conclusion}
In this work, we propose a novel signature-assisted unsupervised bit-vector document representation method, termed \modelname, that allows for fast corpus expansion based on limited relevance judgments. Our work relies on the fact that only a small number of low-frequency terms capture the ``aboutness'' of a document \cite{klein2008revisiting}.  Very common terms that appear in a lot of documents are not as informative of the document topic, while very rare terms do not provide wide coverage of the domain when searching for relevant documents. We leverage such intuition to impose sparsity on the document bit-vectors and reduce the computational burden while ensuring the coverage of the domain-specific long tail distribution.

We conduct experimental analysis with baseline methods and state-of-the-art models. Our results demonstrate the usefulness of our method. More specifically, on a large-scale real-world use-case on terabytes of web data, we find that \modelname reduces the per-document memory footprint by $24\%$, and retrieves $6.8\%$ more rare terms than baselines, on a fraction of time, reducing query execution by $78\%$ w.r.t. the next best strategy. 

The problem of efficiently expanding or creating a document collection through searching petabytes of web documents is relevant to many application domains, from social media monitoring and trends forecasting to epidemics and event extraction. \modelname provides a flexible framework that can be combined with a wide variety of data types. We note that our method is fairly general and can be used with non-document textual queries, \eg a query constructed from a single document, with one or few sentences. In the future, we hope to evaluate our method with different query variants. In addition, we are considering expanding our work in other domains and data, \eg images, video, tables, as well as medical and business intelligence downstream applications, \eg collecting corpora for machine translation, question answering, relation extraction, and taxonomy construction. 

\bibliographystyle{ACM-Reference-Format}
\bibliography{biblio}

\end{document}